\documentclass[sigconf, nonacm]{acmart}
\usepackage{xspace}
\usepackage{listings}
\usepackage{xcolor}
\hypersetup{
    colorlinks=true,
    linkcolor=blue,
    urlcolor=blue,
    citecolor=blue
}
\AtBeginDocument{%
  }

\setcopyright{acmlicensed}
\copyrightyear{2018}
\acmYear{2018}
\acmDOI{XXXXXXX.XXXXXXX}
\acmConference[Conference acronym 'XX]{Make sure to enter the correct
  conference title from your rights confirmation email}{June 03--05,
  2018}{Woodstock, NY}
\acmISBN{978-1-4503-XXXX-X/2018/06}

\begin{document}

\title{Multi-Object Advertisement Creative Generation}
\author{%
Jialu Gao\textsuperscript{1, 2\textdagger},
Mithun Das Gupta\textsuperscript{2},
Qun Li\textsuperscript{2},
Raveena Kshatriya\textsuperscript{2},
Andrew D. Wilson\textsuperscript{3},
Keng-hao Chang\textsuperscript{2},
Balasaravanan Thoravi Kumaravel\textsuperscript{3}\\[0.5ex]
\textsuperscript{1}CMU, USA \quad
\textsuperscript{2}Microsoft AI, USA \quad
\textsuperscript{3}Microsoft Research, USA\\[0.5ex]
\texttt{jialug@andrew.cmu.edu}
\texttt{\{migupta,qul,rkshatriya,kenchan,awilson,bala.kumaravel\}@microsoft.com}\\[0.25ex]
}
\renewcommand{\shortauthors}{Gao et al.}

\newcommand{\ourname}{CreativeAds\xspace}
\newcommand{\participants}{5}

\begin{abstract}
Lifestyle images are photographs that capture environments and objects in everyday settings. In furniture product marketing, advertisers often create lifestyle images containing products to resonate with potential buyers, allowing buyers to visualize how the products fit into their daily lives. While recent advances in Generative Artificial Intelligence~(GenAI) have given rise to realistic image content creation, their application in e-commerce advertising is challenging because high-quality ads must authentically representing the products in realistic scearios. Therefore, manual intervention is usually required for individual generations, making it difficult to scale to larger product catalogs. To understand the challenges faced by advertisers using GenAI to create lifestyle images at scale, we conducted evaluations on ad images generated using state-of-the-art image generation models and identified the major challenges. Based on our findings, we present \ourname, a multi-product ad creation system that supports scalable automated generation with customized parameter adjustment for individual generation. To ensure automated high-quality ad generation, \ourname innovates a pipeline that consists of three modules to address challenges in product pairing, layout generation, and background generation separately. Furthermore, \ourname contains an intuitive user interface to allow users to oversee generation at scale, and it also supports detailed controls on individual generation for user customized adjustments. 
We performed a user study on \ourname and extensive evaluations of the generated images, demonstrating \ourname's ability to create large number of high-quality images at scale for advertisers without requiring expertise in GenAI tools.\\
\begingroup
\renewcommand\thefootnote{}
\footnotetext{\textsuperscript{\textdagger}The work was done during an internship at Microsoft.}
\endgroup

\end{abstract}

\keywords{Generative AI, Image Generation, Diffusion Models}
\begin{teaserfigure}
  \includegraphics[width=0.98\textwidth]{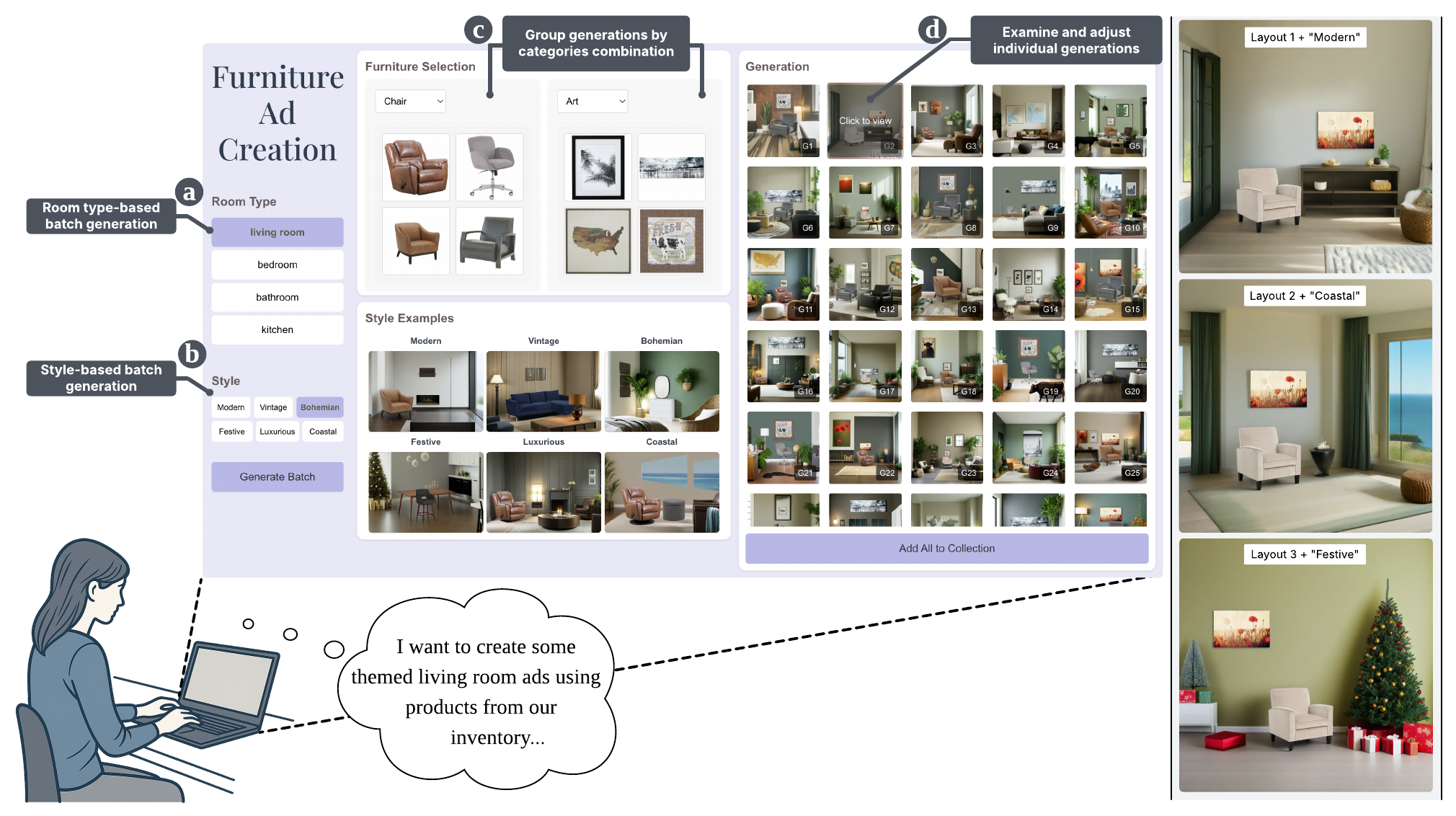}
  \caption{Overview of \ourname, an interactive system that supports scalable high-quality multi-product ad generation.  Supplemental demo video can be found \href{https://youtu.be/LrlF79k1Q28}{here}.}
  \label{fig:teaser}
\end{teaserfigure}

\received{20 February 2007}
\received[revised]{12 March 2009}
\received[accepted]{5 June 2009}

\maketitle

\newpage
\section{Introduction}

In e-commerce, visually appealing advertisement images are crucial for enhancing user engagement and driving purchasing decisions. Studies have demonstrated that promoting products in images of everyday contexts, which are referred to as lifestyle images, significantly increases click-through rates and sale conversion rates~\cite{10.1145/3678610.3678611, mishra2020learningcreatebetterads, wang2021hybridbanditmodelvisual}. However, manually creating high-quality lifestyle images can be expensive and challenging, particularly for vendors managing large product catalogs. As a result, major advertising platforms such as Google, Amazon, Microsoft, and Meta are increasingly investing in automated solutions to generate lifestyle images at scale. These efforts aim to embed generative tools directly into advertiser-facing interfaces, empowering businesses to efficiently generate, review, and customize ad creatives with minimal manual effort—while maintaining high standards of quality, accuracy, and brand integrity. 
 \\

Recent advances in generative AI models have led to the creation of realistic content in various forms, including texts~\cite{openai2024gpt4technicalreport, ouyang2022instructgpt}, images~\cite{ramesh2022hierarchicaltextconditionalimagegeneration, ho2020ddpm, song2022dddim, podell2023sdxl, ldm}, and videos~\cite{ho2022imagenvideo, singer2022makeavideo, text2video-zero}.
In particular, in the field of image generation, Diffusion Models~\cite{ho2020ddpm, song2022dddim, podell2023sdxl, ldm, nichol2021glide, saharia2022photorealistic, dhariwal2021diffusion} have demonstrated an unprecedented ability to produce photorealistic images from text prompts. In addition to text, visual inputs, such as depth maps and edge maps, can serve as additional conditioning factors for image generation~\cite{zhang2023adding}. Image inpainting, a closely related research area, focuses on modifying specific parts of a given image based on textual or visual prompts. Building on these techniques, previous works~\cite{mishra2020learningcreatebetterads, ku2023stagingecommerceproductsonline, yang2024newcreativegenerationpipeline, du2024reliableadvertising, wang2025generateecommerceproductbackground} in e-commerce advertising have leveraged generative models to automatically create compelling lifestyle images by generating realistic and engaging backgrounds for products. \\

While existing methods have shown promising results in generating ad images featuring a single product, they often struggle when extended to multi-product scenarios.  This is largely due to the added complexity of maintaining authenticity—a key requirement in advertisement imagery. Introducing multiple products necessitates careful consideration of product pairings, spatial arrangement, relative sizing, and environmental context to ensure all items are presented naturally and cohesively. Without addressing these challenges, current state-of-the-art methods frequently produce visually unappealing results and compromise authenticity, leading to issues such as unnatural appendages, missing parts, or distorted representations. Such misrepresentations can lead to customer confusion about the products—an outcome that not only concerns advertisers, but also violates strict platform policies that prohibit altering the visual integrity of user-submitted product content. Ensuring product authenticity is therefore not just a matter of quality, but a compliance requirement for platforms that host these ads.\\ %

Despite the challenges, featuring multiple products together in advertisements can be highly beneficial and potentially profitable for businesses. One of the most straightforward benefits is the encouragement of bundled purchases, which simplifies the shopping process for customers. In the furniture industry, for instance, it is common practice to display fully designed room scenes~\cite{shopbyroom-ikea, shopbyroom-amazon, shopbyroom-livingspace, shopbyroom-potterybarn}. Presenting multiple items in a cohesive, styled environment helps customers better visualize how products might look together in their own space. By featuring multiple products in a cohesive and styled environment, customers can better visualize how items will look together in their own space. Moreover, showcasing products in everyday, realistic scenarios fosters a more immersive shopping experience. When users are able to browse interactively—clicking on individual products within a scene—they tend to engage more deeply with the content, which has been shown to increase purchase intent~\cite{product-tagging-pinterest}.\\

In this work, we aim to tackle the problem of automatically generating large volumes of multi-product lifestyle images using commercial furniture product catalogs. These catalogs are often assembled through automated web scraping of business websites or manually uploaded by retailers\footnote{https://support.google.com/merchants/answer/6324350?hl=en}. Unlike curated datasets captured under controlled conditions, catalog product images are usually captured independently—using unknown cameras, lighting setups, viewpoints, and focal lengths—making it difficult to compose them into cohesive scenes.  Moreover, while it is possible to manually create high-quality ads using recent general-purposed image creation methods~\cite{chung2023promptpaint, brade2023promptifytexttoimagegenerationinteractive, wang2024promptcharm}, doing so is labor-intensive, requiring careful artistic judgment and design expertise, and is not scalable to accommodate vast product catalogs. To better inform the design of our framework, we first developed a pipeline leveraging current state-of-the-art general-purpose generative models. We then conducted an expert evaluation of the generated ad images to identify recurring issues and failure modes with two authors of this paper, one with extensive experience in image generation and the other with a background in retail advertising. Using their combined domain expertise, they systematically evaluated the outputs and concluded the observed issues into the following categories:  (1) incompatible product pairings, where unrelated products are combined or where product images captured from significantly different viewpoints are juxtaposed; (2) inaccurate product scaling, where product sizes do not reflect their actual proportions; (3) unrealistic layouts, where product placement appears unnatural or awkward; and (4) background generation artifacts, where the generated environment fails to highlight the products effectively or introduces visual distortions that alter product appearance.  These issues not only degrade the shopping experience, but also risk serious breaches of trust between advertising platforms and their clients. Inaccurate or misleading representations can violate platform policies, damage brand credibility, and expose the system to legal and reputational risks—making the reliable generation of authentic, high-quality images a critical requirement for the adoption of generative models in product advertising.\\

To address these problems, we propose \ourname, a pipeline that can automatically generate batches of high-quality lifestyle images from a large product catalog and allows for user-friendly individual edits and refinements. Firstly, \ourname proposes a 3D-aware approach to pair the products by aligning their orientation obtained by 3D reconstruction. Secondly, to accurately determine the placement and scaling of products, \ourname uses pretrained vision-language model~(VLM) to retrieval product dimensions from the metadata, and generates a realistic layout based on the product images. Finally, \ourname integrates a masked inpainting pipeline based on state-of-the-art diffusion models to achieve high-quality background scene generation. \\

To evaluate the usability and effectiveness of \ourname, we conducted comprehensive experiments across multiple dimensions using furniture data from the publicly available ABO dataset~\cite{collins2022abodataset}. The ABO dataset\footnote{\url{https://amazon-berkeley-objects.s3.amazonaws.com/index.html}} is a dataset of Amazon products where each item is provided with structured metadata corresponding to information that is publicly available on the listing’s
main webpage. We selected this dataset because it closely resembles the types of commercial product catalogs that businesses provide—either via web scraping or manual upload—and is therefore a suitable proxy for real-world deployment scenarios. Using this dataset, we tested \ourname's ability to generate lifestyle images at scale, simulating practical use cases for advertisers. 
To efficiently large amount of generated ads, we perform extensive evaluation using pre-trained models and vision-language models. The quantitative metrics also demonstrate that our images achieve high fidelity and aesthetic scores. To further validate the effectiveness of our design, we performed an ablation study on key modules in \ourname. The results show that  \\

In summary, our main contributions are as follows:
\begin{itemize}
    \item We propose a pipeline that can automatically generate high-quality lifestyle images that showcase multiple products.
    \item We develop a system that enables ad image creation at scale with minimal effort required from users.
\end{itemize}

\section{Relative Work}

\subsection{Diffusion Models for Image Generation}
Text-to-image diffusion models~\cite{ho2020ddpm, song2022dddim, podell2023sdxl, ldm, nichol2021glide, saharia2022photorealistic, dhariwal2021diffusion} are a class of probabilistic generative models that begin with random noise and iteratively refine it by removing noise at each diffusion step, ultimately producing a coherent image. These models often use text prompts to guide the denoising process via techniques like Classifier-Free Guidance~\cite{ho2022classifierfreeguidance}, which steers the generation toward attributes specified in the input text. To enable more fine-grained control, extensions such as ControlNet~\cite{zhang2023adding} incorporate additional image-based inputs—such as depth maps, edge detections, or semantic masks—offering greater flexibility in conditioning the output. Another line of research focuses on layout-conditioned generation~\cite{li2023gligen, NEURIPS2023_3a7f9e48, Xie_2023_ICCV, Couairon_2023_ICCV, bartal2023multidiffusionfusingdiffusionpaths}, where the generation process is guided by spatial constraints, either through modified training or test-time sampling techniques. Beyond image generation, diffusion models have also been widely explored for image editing tasks~\cite{Cao_2023_ICCV, brooks2023instructpix2pixlearningfollowimage, hertz2022prompttopromptimageeditingcross, kawar2023imagictextbasedrealimage, wei2024omniedit}. These approaches take an input image and modify it based on text prompts, enabling intuitive and flexible manipulation of existing visual content. A specialized application within image editing is image inpainting~\cite{podell2023sdxl, zhuang2024taskworthwordlearning, lugmayr2022repaint, ju2024brushnetplugandplayimageinpainting}, where the objective is to modify or complete specific regions of an image while maintaining overall visual coherence. These models are particularly effective at reconstructing missing areas or seamlessly integrating new content into existing scenes.

\subsection{Generated AI Powered Design} Generative models have been increasingly integrated into design workflows across a wide range of domains, including fashion~\cite{FashionFuture, StyleMe}, landscape rendering~\cite{PlantoGraphy}, 3D design~\cite{chen2024memovis}, and user interface design~\cite{stylingtheweb}. In the context of general-purpose image creation, recent systems such as PromptCharm\cite{wang2024promptcharm}, Promptify\cite{brade2023promptifytexttoimagegenerationinteractive}, PromptPaint\cite{chung2023promptpaint}, MagicQuill\cite{liu2025magicquillintelligentinteractiveimage}, and PromptMap\cite{10.1145/3708359.3712150} have introduced interactive interfaces that help users more effectively express their intent for image generation. These tools explore novel interaction paradigms aimed at streamlining the creative process and lowering the barrier to entry, allowing users with diverse levels of expertise to engage in expressive visual content creation. Furthermore, the paper "We Are Visual Thinkers, Not Verbal Thinkers!"\cite{visualthinkers} offers valuable insights into the design of future generative AI tools, emphasizing the importance of multi-modal input methods, intuitive prompt control, and support for iterative exploration—principles that have directly inspired the design of \ourname. In the context of e-commerce advertising, Zhou et al. \cite{Zhou_2020} leveraged visual question answering models to infer advertising themes based on ad creatives (text and images). With advances in image content creation, inpainting models have emerged as tools for crafting visually appealing ad images. Ku et al. \cite{ku2023stagingecommerceproductsonline} proposed a retrieval-based copy-paste method, combined with generative adversarial networks, to achieve high-quality inpainting of product images with solid backgrounds. More recently, diffusion models \cite{yang2024newcreativegenerationpipeline, du2024reliableadvertising, wang2025generateecommerceproductbackground} have been employed to generate attractive and natural contextual backgrounds for product images with plain backgrounds. However, all of the above mentioned approaches are limited to creating ad images showcasing a \textit{single product} and do not account for related or complementary products. In this work, we address this gap by introducing a novel pipeline for creating high-quality ad images that feature multiple products.

\section{Formative Study} 

\begin{figure}[h]
  \centering
  \includegraphics[width=\linewidth]{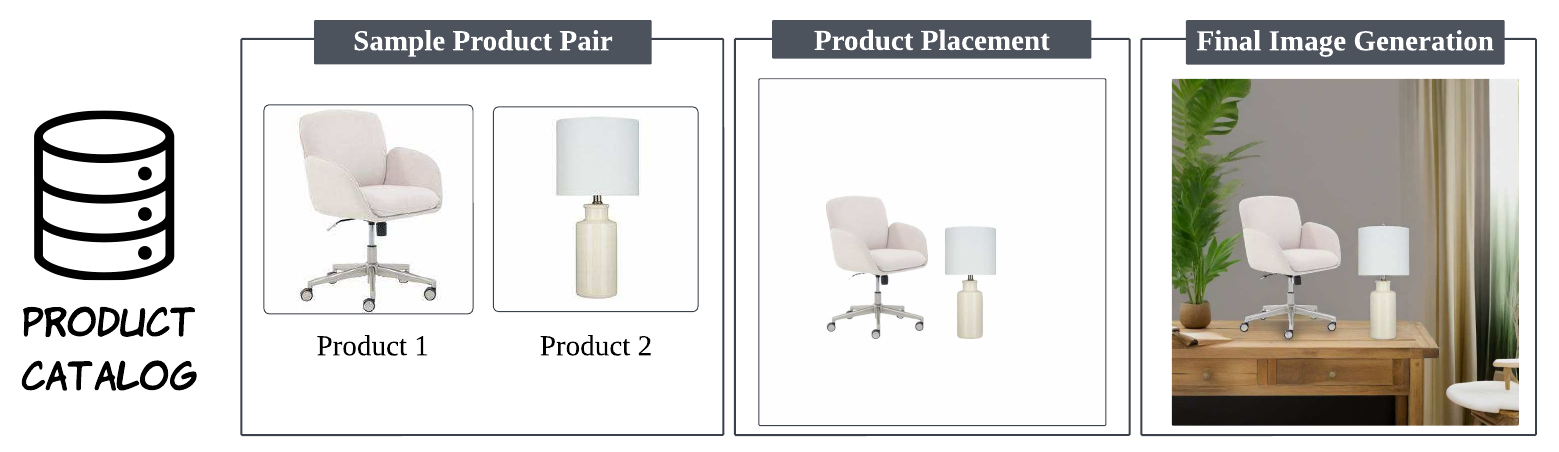}
  \caption{The initial workflow design for multi-product ad generations consists of three stages: product sampling, placement, and background generation.}
  \label{fig:workflow}
\end{figure}

To better understand the challenges of using existing AI models for high-quality ad images for users, we implement a simple workflow using state-of-the-art general purpose generative models for multi-product furniture ad creation, as illustrated in Fig.~\ref{fig:workflow}. There are three stages in the workflow: 1) Product pairing, where the workflow randomly samples a pair of products from the full catalog. 2) Product placement, where the workflow puts the products side-by-side on a blank canvas. 3) Final generation, where the workflow generates the appropriate background for the canvas. \\

We randomly sampled 200 pairs of furniture data from the publicly available Amazon Berkeley Object benchmark~\cite{collins2022abodataset} for the test.  The pipeline for product placement and background generation is built using ComfyUI\footnote{https://github.com/comfyanonymous/ComfyUI}. The placement is implemented through prompting GPT-4o~\cite{openai2024gpt4technicalreport} to generate coordinates of each product on a square canvas with side-by-side placement. The background is generated using the Stable Diffusion XL Inpainting model~\cite{podell2023sdxl}, guided by a simple text prompt template: ``a {product 1} and {product 2} in a room'', where the placeholders are filled with the category labels of the respective products.

\subsection{Problems with Existing Methods}
\begin{figure}[h]
  \centering
  \includegraphics[width=\linewidth]{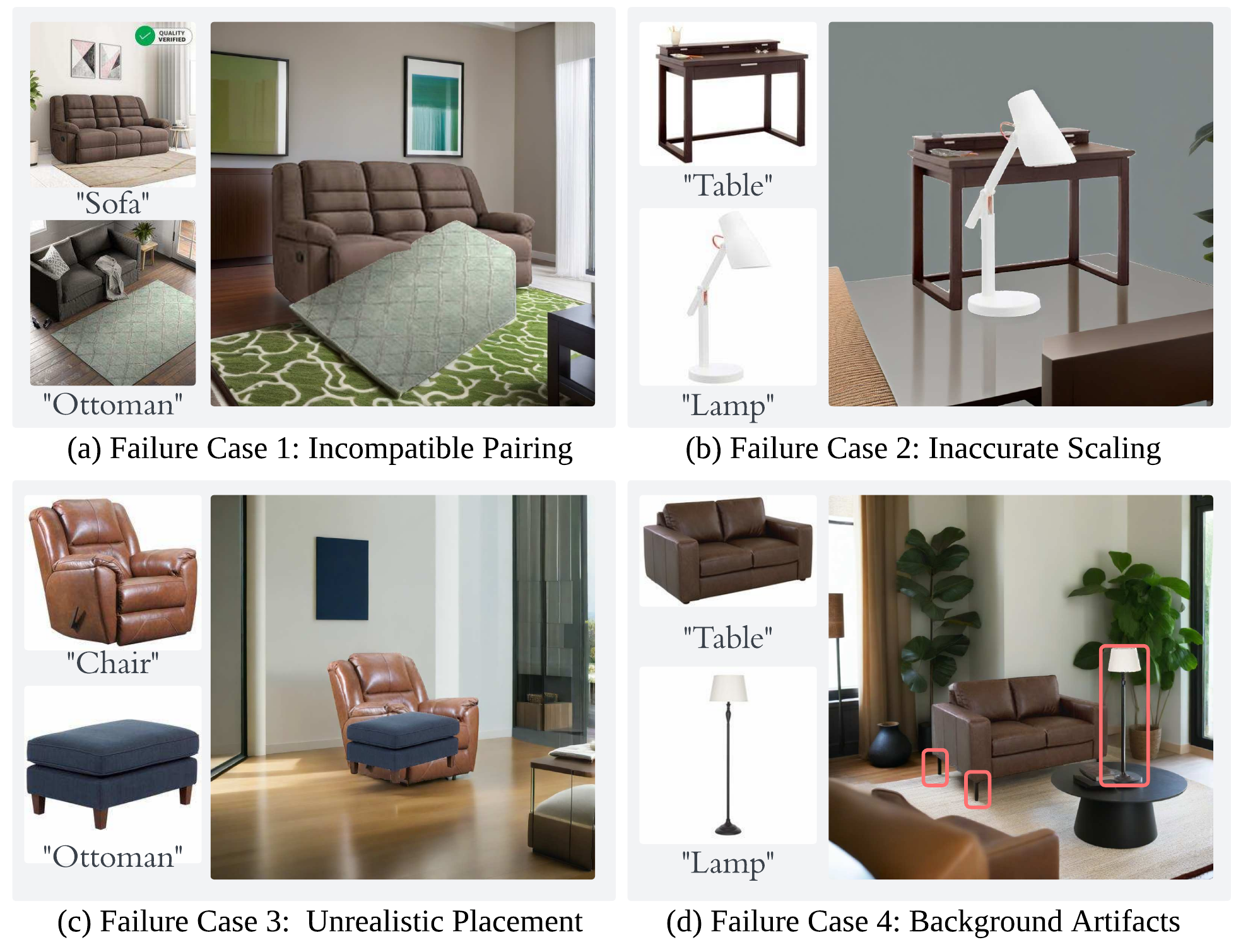}
  \caption{Common failure cases for directly applying state-of-the-art generation models for multi-product ad creations.}
  \label{fig:failure-case}
\end{figure}

We then asked two authors of this paper to label the generated results—one with deep expertise in text-to-image and image-to-image generation techniques, and the other with a professional background in advertising and creative workflows. The labeling was conducted following a protocol inspired by advertising experts: Given a generation consists of $($\textit{product\ 1}, \textit{product\ 2}, \textit{generated\ ad}, \textit{generation\ prompt}$)$, users are asked to select one of all labels that best reflect the generation: 
\begin{itemize}
    \item [0.]The generated ad perfectly showcases the two products according to the prompt.
    \item [1.]The product images are not matched with their category labels
    \item [2.]The products pairing is not natural 
    \item [3.]The relative sizes of products is not accurate
    \item [4.]The products are not placed at reasonable locations
    \item [5.]The visual appearance of the two products are changed in the generated ad
    \item [6.]The background being generated are not featuring well the functionality of the products
\end{itemize}

Results showed that on average, only $7.5\%$ of the generated ads are considered high-quality. The authors analyzed failure cases and systematically categorized common issues into four primary classes, as shown in Fig.~\ref{fig:failure-case}. The first type of failure occurs when incompatible products are paired together. One example is illustrated in Fig.~\ref{fig:failure-case}~(a), where the product images are not taken from a similar view point, and thus it will be impossible to create a coherent ad image containing the products without modifying the product images. The second failure case is where the sizes of product images are not reflecting the actual product sizes~(Fig.~\ref{fig:failure-case}~(b)). This not only results in generations that are not photo-realistic that fails to attract users, but will also lead to user misunderstandings about the actual information of the products. The third case is where the model generates unsatisfying results is placing the products in an unnatural way, such as stacking the products together in the center of the canvas, as shown in Fig.~\ref{fig:failure-case}~(c). The last failure case is the artifacts brought by the generated background, an example showing in Fig.~\ref{fig:failure-case}~(d). This includes occasions when the product appearance is changed, like adding a leg to the sofa chair. It also includes generating a background with poor affordance with the foreground, such as the lamp is sitting on top of a coffee table.

Additionally, we collected feedback from the authors who labeled the results. One observation was that it is time-consuming for users to holistically evaluate the generated images by manually opening them one by one—even with just 200 images in our pilot study. Scaling to thousands of images would require more efficient tools to support visual exploration and assessment. Another key insight was that, for most failure cases, having human controls would be highly beneficial, allowing users to intervene and guide the generation process when needed.

\subsection{Design Considerations}
\label{sec:3-2}
Our goal is to enable users to automatically generate a large volume of high-quality ad images from extensive product catalogs. These images should showcase multiple complementary products arranged in realistic, everyday settings that are visually appealing to users. Despite the scale of generation, users should be able to easily browse, select, and customize individual outputs according to their preferences. Based on the summary of the preliminary study, we formulate the following design considerations for our scalable GenAI ad creation tool.

\begin{itemize}
    \item \textbf{D1: Support generation from noisy product catalog.} The tool should be capable of handling large volumes of noisy, unstructured product data often found in real-world business catalogs. It must organize and preprocess this raw input—such as inconsistent naming, missing attributes, or redundant entries—into a structured format that enables efficient product pairing and reliable image generation.
    \item \textbf{D2: Address the limitations of general image generative models for e-commerce furniture ad generation.} The tool should contain an image generation pipeline specific designed for advertising, integrating various techniques to address common shortcomings of general-purpose models—particularly in preserving product authenticity. By tailoring the generation process to the specific needs of furniture advertising, the system can consistently produce high-quality outputs across a diverse range of product samples, significantly improving the overall yield of usable ad images.
    \item \textbf{D3: Optimize between scalable automation and individual refinement.} The tool should incorporate an adaptive refinement workflow, where users can set broad parameters for batch generation and selectively enhance outputs with guided interventions. While on average the pipeline can handle most inputs, there are corner cases where the method fails to create acceptable images. In such cases, the tool should allow users to correct these errors in a cost-effective manner.
    \item \textbf{D4: Allow user-friendly examination and review for large amount of generated ads.} Given the scale of generation, potentially thousands of ad images, it is critical to support users in efficiently browsing, evaluating, and managing the outputs. The system should provide intuitive interfaces (e.g., grid layouts, clustering by visual similarity or product types) for overview to circumvent manually inspecting each image one by one. 
\end{itemize}

\section{\ourname: System Design}
\begin{figure*}[h]
  \centering
  \includegraphics[width=\linewidth]{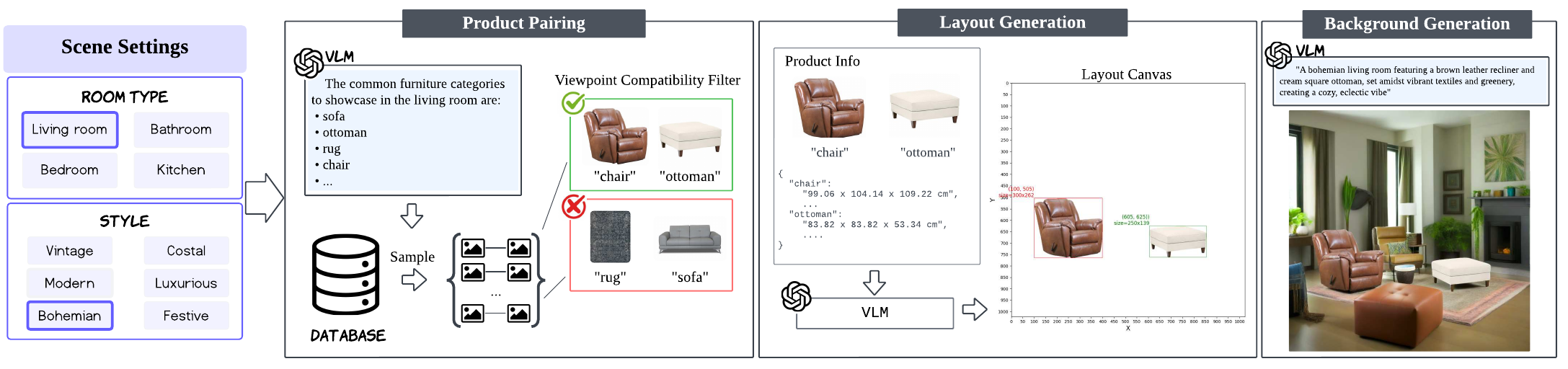}
  \caption{Overview of the system workflow. \ourname can batch generate high-quality ads given a user specified room-type and generation style. The system consists of three core modules, each designed to address a key challenge in realistic and authentic ad creation. First, the Product Pairing Module selects product combinations that are both semantically and visually compatible, ensuring functional coherence and viewpoint consistency. Next, the Layout Generation Module determines the spatial arrangement and relative sizes of products on a blank canvas to achieve a natural and balanced composition. Finally, the Background Generation Module inpaints the surrounding environment to highlight the selected products while preserving overall visual authenticity.}
  \label{fig:main-figure}
\end{figure*}
\label{sec:4-1}
This section introduces \ourname, a tool that enables scalable ad generation for large online product catalogs with customized user controls. We start with an overview of the system~(Sect.~\ref{sec:4-1}) and how it aligns with the design considerations summarized in Sect.~\ref{sec:3-2}. Subsequently, we introduce the interface design and workflow in Sect.~\ref{sec:4-2}. Then we describe the three modules in \ourname that make high-quality ad image generation possible: product pairing module~(Sect.\ref{sec:4-3}), layout generation module~(Sect.\ref{sec:4-4}), and background generation module~(Sect.\ref{sec:4-5}). Finally, we provide the implementation details in Sect.~\ref{sec:4-6} for all algorithms used in \ourname.

\subsection{System Overview}
\ourname is an interactive tool that enables users to generate visually compelling lifestyle advertisement images at scale by leveraging a large online product catalog. While users may have prior experience with image design, we do not assume familiarity with recent genAI models or workflows. Even for users proficient in GenAI, manually managing the creation of thousands of images is impractical. To facilitate scalable and high-quality ad generation, we introduce three key modules designed to automate the process and support diverse input data, as illustrated in Fig.\ref{fig:main-figure}:
\begin{itemize}
    \item \textbf{Product pairing module.} The online product catalog contains a wide variety of everyday objects, many of which do not naturally co-occur in the same physical environment—such as sofas and showers, or toilets and dining tables. Additionally, product images are captured using different cameras and from varying angles. For instance, rugs are typically photographed from a top-down perspective to maximize visual clarity, whereas chairs and beds are often captured from a front-facing angle. As a result, not all product combinations from this extensive catalog are visually compatible for composing realistic ad images. To support scalable generation with large product catalogs ~(\textbf{D1}) and to reduce failure cases caused by implausible pairings~(\textbf{D2}), we introduce a Product Pairing Module. This module pre-processes raw product data to ensure semantically and visually compatible pairings. It leverages a vision-language model (VLM) to extract structured product metadata from catalog entries and employs a metric-depth estimation model to infer camera viewpoints, filtering out incompatible object combinations accordingly. 
    \item \textbf{Layout generation module.} From the exploratory study, we notice that two of the four common failure cases for multi-product ad creation result from the layout: wrong relative scaling of the products and unrealistic product placement. To overcome these challenges and improve the percentage of high-quality images being generated~(\textbf{D2}), we design the layout generation module to help. Instead of directly generating the layout, the module first uses a VLM agent to set some constraints: it retrieves product dimension information from the metadata obtained from the product catalog and generates a textual description for the ideal product placement by analyzing the two product images. Then, the retrieved dimension and layout description are used as prompting conditions for a VLM to generate the layout by specifying coordinates and sizes of products on a fixed-size canvas. By doing so, we are breaking down the complex reasoning happening inside the VLM during layout generation into two steps, allowing users to step in and intervene with the generation when things are going wrong, balancing between automated generation and individual refinement~(\textbf{D3}). 
    \item \textbf{Background generation module.} The ultimate goal of \ourname is to create visually appealing ad images while accurately showing the products exactly as they are~(\textbf{D2}). This is where the background generation module comes in. To preserve the product details, we introduce a segmentation model to mask out the products before generation, ensuring that only the empty space in the canvas is changed. When creating attractive backgrounds for products, we also break it down into two parts, where a VLM is first asked to generate a background description prompt given the canvas obtained from the layout generation module, and subsequently an inpainting model is used to inpaint the background with the prompt. This will support both the scalable automated generation, and individual adjustment where users can tweak the background prompt as they like~(\textbf{D3}). To better minimize the failure case where the generated background modifies the products~(\textbf{D2}), we incorporate ControlNet~\cite{zhang2023adding} as additional structural constraints to the background generation.
    
\end{itemize}

\subsection{User Interface}
\label{sec:4-2}
\begin{figure*}[h]
  \centering
  \includegraphics[width=\linewidth]{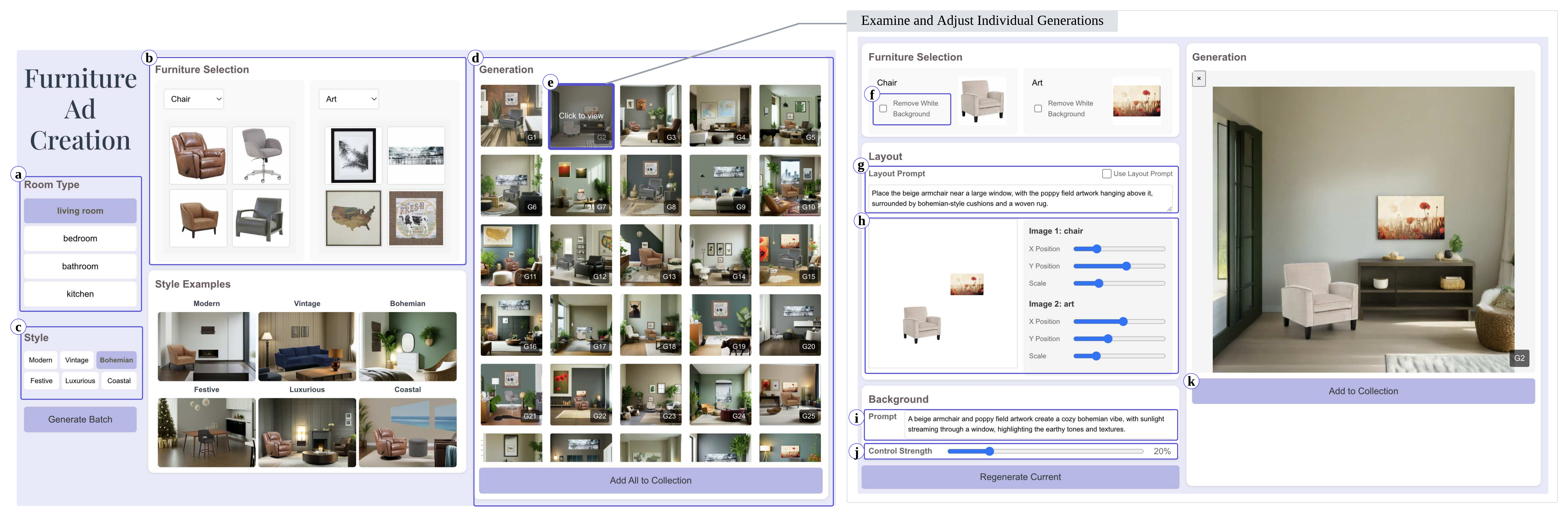}
  \caption{User Interface of \ourname. The interface of \ourname is designed to support efficient control and inspection of batch ad generation. On the left panel, users can specify high-level parameters such as the desired room type and visual style. The central panel allows users to select specific product category combinations to guide the generation process. On the right panel, a gallery view presents the generated images in a compact, scrollable layout. Users can click on individual thumbnails to view full-size images, inspect generation details, and make fine-grained adjustments to individual outputs as needed.}
  \label{fig:user-interface}
\end{figure*}

In this section, we present a description of the \ourname user interface. An illustration of the interface is shown in Fig.\ref{fig:user-interface}. The main interface enables users to perform batch generations in groups, providing a streamlined way for users to browse and evaluate the completed ad images~(\textbf{D4}). It also allows them to inspect and refine individual generations as needed. \\

In Fig.\ref{fig:user-interface}, we illustrate how the interface is designed to support automated batch generation as well as fine-grained control for each individual ad creation. The interface consists of the following three panels: 
\begin{itemize}
    \item \textbf{Left panel.} The left panel allows users to configure parameters for batch generation. Given that the tool can generate thousands of images simultaneously—making them difficult to review all at once—we organize the generations by room type and generation style. Users can specify the room type that the lifestyle image should depict~(Fig.\ref{fig:user-interface}(a)) and select from a variety of generation styles to guide the visual outcome~(Fig.\ref{fig:user-interface}(c)). 
    \item \textbf{Central panel.} The central panel is to support more detailed and fine-grained control for users. Firstly, in the ``Furniture Selection'' section~(Fig.\ref{fig:user-interface}(b)), the interface will display a list of possible furniture categories retrieved from the catalog according to the specified room type. Users can then use the drop-down list to choose a pair of furniture categories to be advertised. Once the user has selected certain categories, a grid of sampled product images is displayed below to better inform users what each category looks like. Secondly, the central panel also serves as a playground for users to modify a certain individual generation in detail. When users identify an image from the gallery that they are not satisfied with, they can click on the image~(Fig.\ref{fig:user-interface}(e)) to enter the review and adjust interface. In rare cases, the product backgrounds in the example image are copied into the final ads, which is not desirable. By checking the ``Remove White Background''~(Fig.\ref{fig:user-interface}(f)), users can remove the white spaces from the product's example image to allow better segmentation. The interface also allows users to change the layout through editing the layout prompt~(Fig.\ref{fig:user-interface}(g)) or directly setting the coordinates and sizes of the products~(Fig.\ref{fig:user-interface}(h)). For the background of the generated ads, users can modify it by changing the background description prompt~(Fig.\ref{fig:user-interface}(i)).  Additionally, users can also control how closely the final image adheres to the layout defined in the canvas by adjusting the control strength slider~(Fig.\ref{fig:user-interface}(j)). A value of $0\%$ indicates no structural constraint is applied, allowing for more freedom in image composition, while a value of $100\%$ enforces strict adherence to the layout—resulting in outputs that are structurally identical to the generated canvas, typically with only plain backgrounds added.
    \item \textbf{Right panel.} The right panel is to display the generation results. After uses click the ``Generate Batch'' button in the left panel, an image gallery will be displayed on the right panel of all lifestyle ads created under the current settings. Users can click on each image in the gallery to enter the adjustment page, where the right panel displays that particular generation in higher resolution to let the user have a clearer view at the image, for better evaluation and customization.
\end{itemize}

\subsection{Product Pairing Module}
\label{sec:4-3}
\begin{figure*}[h]
  \centering
  \includegraphics[width=0.9\linewidth]{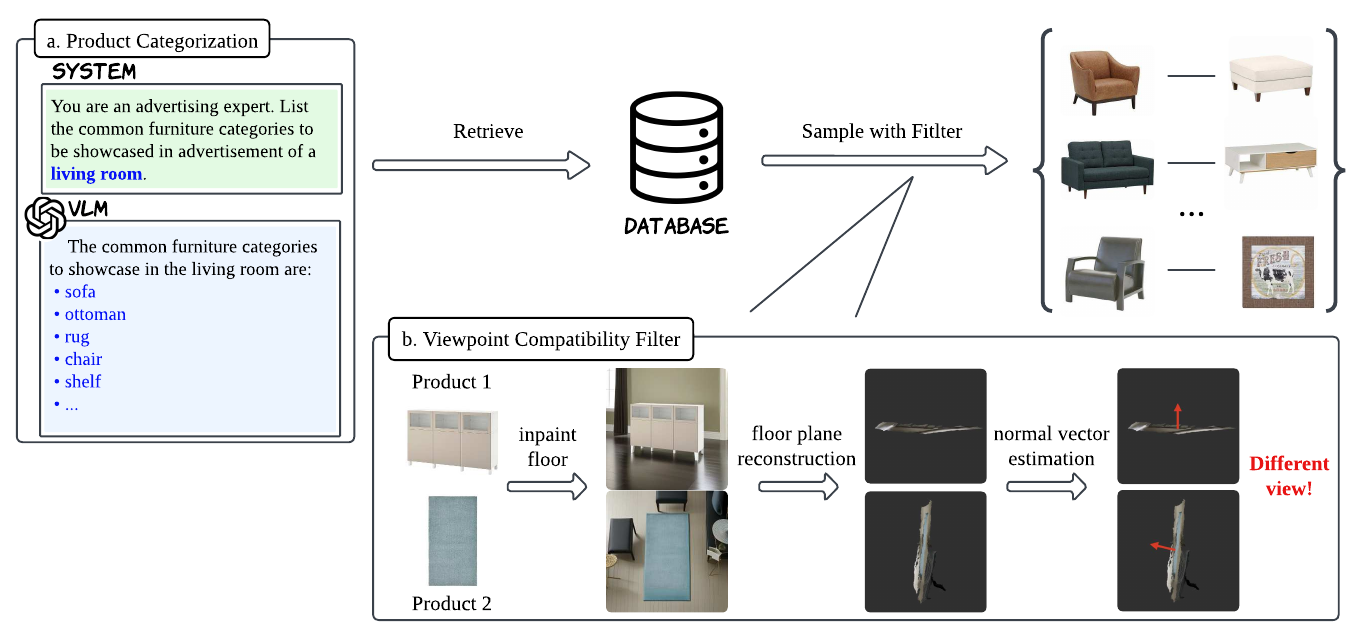}
  \caption{Illustration of the Product Pairing Module. This module first categorizes products into room types using a vision-language model (VLM) and then pairs semantically compatible items. To ensure visual coherence, it further filters product pairs by inferring the camera tilt from product images, selecting only those captured from similar viewpoints.}
  \label{fig:product-pairing}
\end{figure*}

Generating natural and complementary product pairing is an important prerequisite for creating high-quality ad images. A significant challenge of directly using product images from catalogs to create advertisement images is that these images are not taken from canonical sources, possibly captured from unknown, and very different viewpoints, lighting, distance, focal length, etc. To address the challenges of dealing with massive product catalogs where not all of the products are compatible, we introduce the following two processes in product pairing, as illustrated in Fig.\ref{fig:product-pairing}.\\

\noindent \textbf{Product categorization with room types.} To avoid pairing products that typically do not appear together in real-life scenarios—such as placing a dining table in a bedroom—we leverage a VLM to classify all product categories into one of four common household room types: living room, bedroom, kitchen, and bathroom. During automated batch ad creation, products are sampled only from categories associated with the same room type. This approach ensures that all paired products are functionally compatible and contextually appropriate. \\

\begin{figure}[h]
  \centering
  \includegraphics[width=\linewidth]{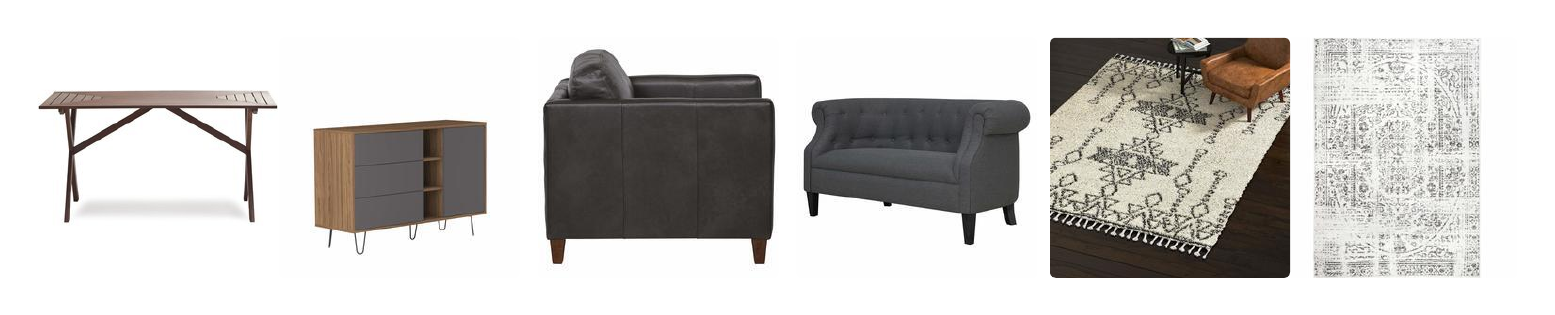}
  \caption{Examples of diverse product images from the catalog. These images are captured using different cameras, lighting conditions, and viewpoints. }
  \label{fig:diverse-products}
\end{figure}
\noindent \textbf{Viewpoint compatibility filter.} In advertising, maintaining product authenticity is a critical requirement, particularly in generated content. To uphold this standard, we restrict our system to using only the provided product images for ad creation, as novel-view synthesis techniques—while visually compelling—can alter the appearance of products in ways that compromise fidelity. However, product catalogs typically contain images captured in varied environments, using different cameras, lighting conditions, and viewpoints (see Fig.\ref{fig:diverse-products}). For each product, we apply 3D reconstruction techniques to estimate the floor plane it rests on and infer the camera's tilt via the normal vector of that plane. By comparing the angular difference between these tilt vectors, we assess viewpoint similarity and filter out mismatched product pairs accordingly.\\

\subsection{Layout Generation Module}
\label{sec:4-4}
\begin{figure*}[h]
  \centering
  \includegraphics[width=0.9\linewidth]{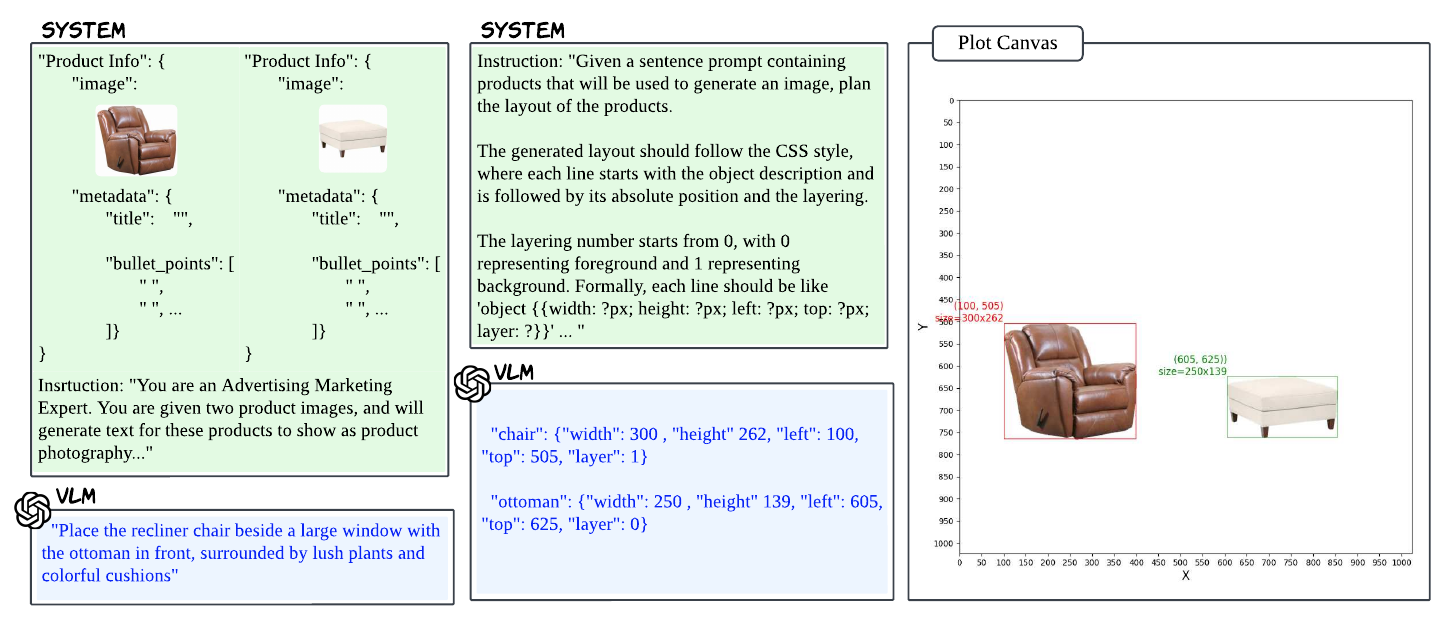}
  \caption{Illustration of the Layout Generation Module. This module first generates a structured layout description prompt and extracts product dimension information from metadata to ground the layout in physical realism. It then leverages a vision-language model (VLM) to predict the spatial coordinates, relative sizes, and layering order of each product within the scene.}
  \label{fig:layout-generation}
\end{figure*}

The layout generation module, which predicts the locations and the relative sizes of the products %
relative to each other, is an extremely challenging module, since its output greatly affects the visual aesthetics of the final image. In \ourname, we design a new workflow that utilizes VLMs to achieve high-quality layout generation, as illustrated in Fig.~\ref{fig:layout-generation}.\\

\noindent \textbf{Layout description prompting.} Since the task of generating a reasonable and realistic layout from scratch is difficult, we break down the problem into two intermediate steps. First, we want to ground the layout design using language as a source of guidance and constraint. Given the product images and metadata information, we first prompt the VLM to generate a text description of the layout containing the two products, such as ``Place the recliner chair beside a large window with
the ottoman in front, surrounded by lush plants and
colorful cushions''. This information will be used to prompt VLMs better to generate the final layout canvas, and will also enable the user to intervene by using language to modify the layout. \\

\noindent \textbf{Layout canvas generation.} After generating a layout description, we will ask VLM to finalize the layout by generating the coordinates and sizes of the products. To make sure the products have accurate relative scaling, we utilize the metadata from the product catalog, which is structured text scraped from the product listing's main webpage, and ask the VLM to extract dimension related information. Subsequently, we input this information along with the layout description into the VLM and obtain the layout. Since multiple products may also lead to occlusion and layering problems, we also explicitly ask the model to return a layer order of the products. Finally, following the coordinates, the sizes, and the layering, we will be able to render a blank canvas containing reasonable locations and scaling of the products. \\

\subsection{Background Generation Module}
\label{sec:4-5}
\begin{figure*}[h]
  \centering
  \includegraphics[width=\linewidth]{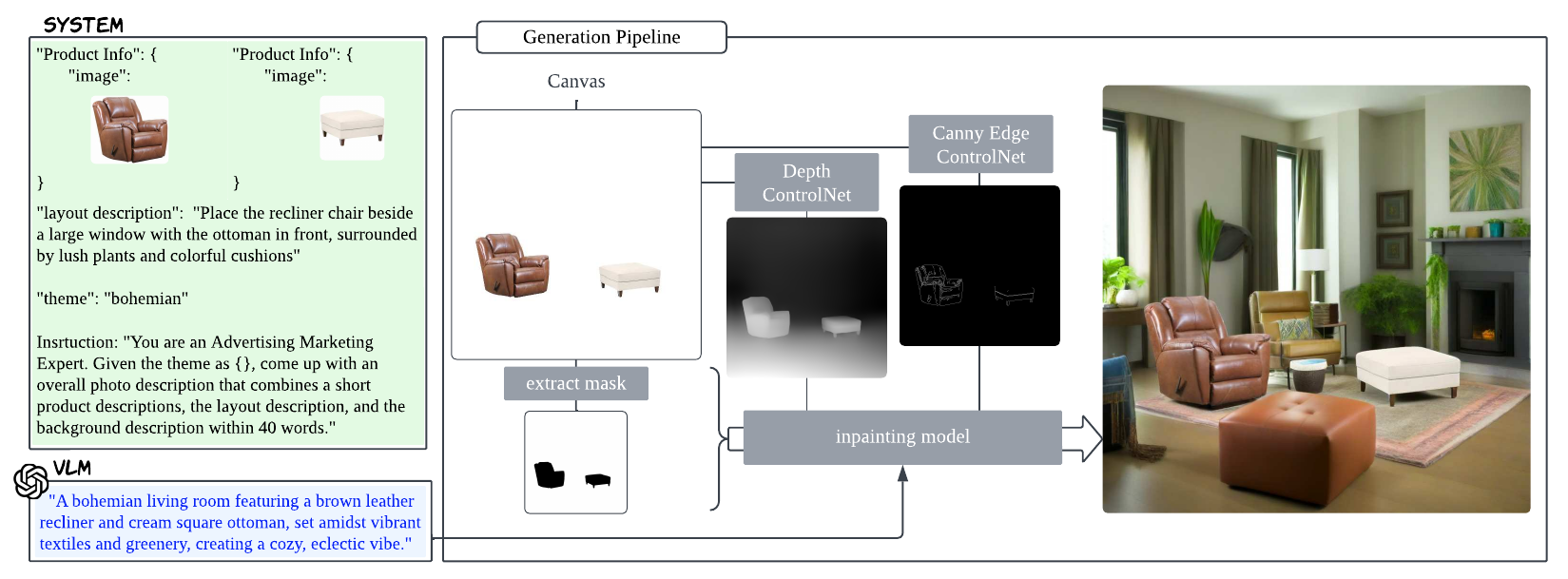}
  \caption{Illustration of the Background Generation Module. This module first employs a vision-language model (VLM) to generate semantically meaningful scene descriptions that guide background synthesis. It then applies masked inpainting with ControlNet to produce realistic, context-aware environments that both highlight the products and preserve their visual authenticity.}
  \label{fig:background-generation}
\end{figure*}

The goal of background generation is to transform the layout canvas with blank background from the layout generation module into a visually attractive, aesthetically pleasing advertisement. It should also follow the restriction that none of the product's appearance can be modified. To this end, we design the background generation module as shown in Fig.\ref{fig:background-generation}. \\

\noindent \textbf{Background description prompting.} For text-to-image diffusion models, the quality of the generated image is highly dependent on the text prompt. Therefore, we adopt VLMs to automatically generate text descriptions for the background, and we pre-define a set of themes for the lifestyle images to encourage VLMs to come up with more diverse and detailed prompt. \\

\noindent \textbf{Masked inpainting.} To satisfy the requirement of no visual changes to the products, we use inpainting models with foreground mask to ensure that only the background is being modified. A segmentation model is used to segment the foreground objects on the canvas and thus obtain the background mask, which will be passed into the inpainting model. During inpainting, the mask will be used to blend the foreground objects and the generated background together, leading to a coherent scene generation. \\

\noindent \textbf{Structural guidance with ControlNet.} In some cases, only masked inpainting is not enough to maintain the product appearance. The inpainted backgrounds can add extensions to the products, such as adding legs to the brown recliner chair in Fig.\ref{fig:background-generation}. To eliminate this kind of error, we adopt ControlNet to enforce the background generation to follow the structure of the input canvas. With a proper level of control strength, the inpainting model can achieve a balance in preserving the structure of the input canvas while adding realistic and semantically meaningful elements to the background. \\

\subsection{Implementation Details}
\label{sec:4-6}
We implemented \ourname's user interface using React as frontend. The backend is built using Flask\footnote{https://flask.palletsprojects.com/en/stable} and ComfyUI\footnote{https://github.com/comfyanonymous/ComfyUI}. The backend is hosted on a Nvidia A100 GPU with 80G of memory. A single generation of an image takes 10 seconds on average. \\

\textbf{Product pairing module.}  We use Metric3D~\cite{metric3d} as our metric depth estimation model for 3D reconstruction and Segment Anything~\cite{sam} for floor plane segmentation. For product information retrieval, we are using GPT-4o~\cite{openai2024gpt4technicalreport} as the VLM agent.\\

\textbf{Layout generation module.} GPT-4o~\cite{openai2024gpt4technicalreport} is used for layout generation. Detailed prompts are provided in the Appendix.\\

\textbf{Background generation module.} For inpainting, we are using checkpoint version 6 from Juggernaut as the base model, with Fooocus Inpaint workflow\footnote{https://github.com/lllyasviel/Fooocus}. Segment Anything~\cite{sam} is used to obtain the foregound and background mask for inpainting. The ControlNet models we used for structural control are the ControlNet Depth\footnote{https://huggingface.co/diffusers/controlnet-depth-sdxl-1.0} and Canny\footnote{https://huggingface.co/diffusers/controlnet-canny-sdxl-1.0} model, with a default strength of 0.2.

\section{Evaluation}
\subsection{Scenario Evaluation}
In this section, we will be presenting a scenario evaluation of \ourname. Since our tool targets generating thousands of images, at this scale the generation may take up to a few days to finish. Therefore, we are providing a scenario walkthrough of our tool to justify it's usability~\cite{10.1145/1294211.1294256}.

\begin{figure*}[h]
  \centering
  \includegraphics[width=0.98\linewidth]{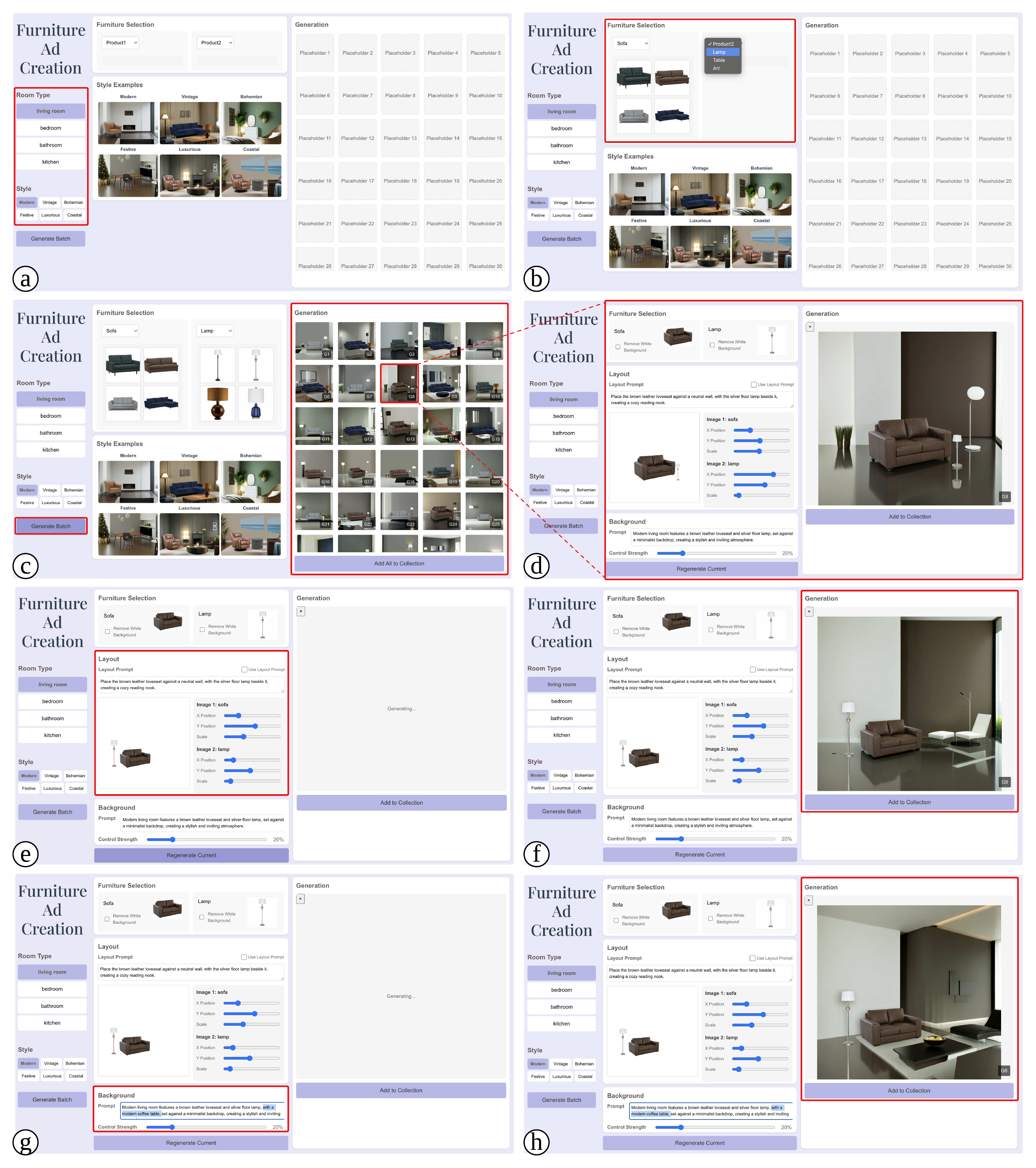}
  \caption{Scenario walkthrough 1 - Monitoring batch generation and adjusting individual ones. \ourname enables users to oversee batch generations organized by room types and generation styles. Additionally, it provides options for users to make adjustments to individual generations based on their specific preferences.}
  \label{fig:scenario-1}
\end{figure*}
Lucy is a small business advertiser who wants to create themed living room ads using products from their inventory. After providing the product catalog to our system, she will be led to the main user interface to supervise batch generation and adjust individual ones.

\noindent \textbf{Room type and style selection.} Lucy will first be asked to select the room type in Fig.\ref{fig:scenario-1}(a), which will be ``living room'' for now. Lucy will also select a generation style from the provided list, for example, ``Modern''. Reference images of each style are shown next to the style selection bar. The room type and style will act as the broader parameters that control all subsequent generations. \\

\noindent \textbf{Category selection.} As soon as Lucy selects the room type, the drop down list in Fig.\ref{fig:scenario-1}(b) will be updated accordingly. When Lucy clicks on the drop down bar, a list of possible product categories to be featured in ``living room'' will be shown. After a category is selected, several sampled images from that category will be displayed below for reference. After setting both categories, Lucy clicks on the ``Generate Batch'' button highlighted in Fig.\ref{fig:scenario-1}(c) to perform the batch generation. This batch generation will automatically sample product pairs according to the categories that Lucy has chosen, and generate ad images for ``Living room'' following ``Modern'' style. \\

\noindent \textbf{Batch generation and review.} When batch generation is finished, all generated ads will be shown on the right panel in the form of an image gallery~(Fig.\ref{fig:scenario-1}(c)). Each thumbnail image in the gallery is clickable to view in details. After browsing through the images, Lucy decides that the 8th image~(denoted as ``G8'') from the batch generation needs a closer look, and clicks on the image and enters the individual generation page, as shown in Fig.\ref{fig:scenario-1}(d).\\

\noindent \textbf{Individual adjustments: Layout.} After taking a closer look at ``G8'', Lucy decides that the layout needs adjusting because the relative scaling of the lamp and sofa is off. Lucy chooses to adjust the layout directly by dragging the slide bar in Fig.\ref{fig:scenario-1}(e) for a different layout, and the modified layout is being shown on the canvas. Then Lucy clicks ``Regenerate Current'' in Fig.\ref{fig:scenario-1}(e) and the interface will be generating a new ad image according to the current layout canvas. When the generation is finished, it will be displayed on the right panel, as shown in Fig.\ref{fig:scenario-1}(f).\\

\noindent \textbf{Individual adjustments: Background.} After examining the new generated ad, Lucy is satisfied with the layout but feels the scene is a bit empty and wants to add additional details to the background. Lucy will go to the Background panel highlighted in Fig.\ref{fig:scenario-1}(g), and edits the generation prompt to add ``a modern coffee table'' to the scene. After making the edits and clicks ``Regenerate Current'' to apply the changes in the background, Lucy will be able to see the newer ad in the right panel~(Fig.\ref{fig:scenario-1}(h)), where a coffee table is being added and it fits the overall style. Lucy decides that this is a qualified ad and adds it to the collection by replacing the older version through a click of the ``Add to Collection'' button in Fig.\ref{fig:scenario-1}(h).\\

\begin{figure*}[h]
  \centering
  \includegraphics[width=0.8\linewidth]{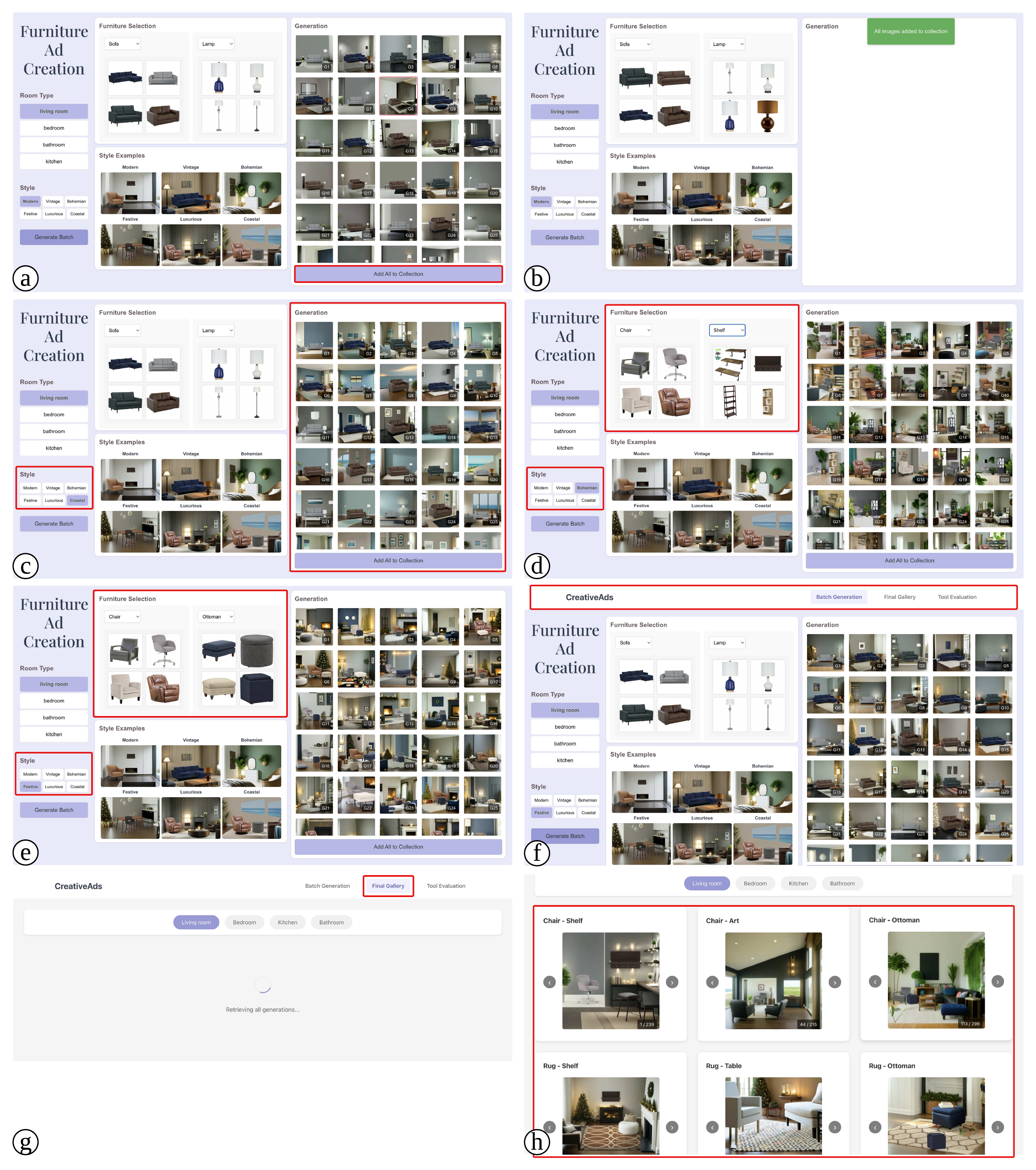}
  \caption{Scenario walkthrough 2 - Review. \ourname groups the generated images by room types, styles, and product category pairs. The results are displayed in a carousel for easy browsing, allowing for an efficient final review of the generated content.}
  \label{fig:scenario-2}
\end{figure*}

\noindent \textbf{Review of different styles.} After browsing through the generations in the ``Gallery'' and making adjustments as needed, Lucy is satisfied with the overall quality and wants to generate more living room ads containing ``Sofa'' and ``Lamp'' but under different themes. Lucy will first save all the modification made before by clicking ``Add All to Collection'' in Fig.\ref{fig:scenario-2}(a), and the system will record all the changes under the current user in the disk storage~(Fig.\ref{fig:scenario-2}(b)). Next, Lucy can click on other styles in the left panel, such as ``Coastal'', and perform batch generation. When results are ready, they will be shown in under ``Gallery''~(Fig.\ref{fig:scenario-2}(c)).

\noindent \textbf{Review of different product pairings.} Apart from the category combinations of ``Sofa'' and ``Lamp'', there are other possible furniture category combinations that are suitable for living room. Similarly, Lucy can click on the drop down list again and set another pair of product categories to be featured in ads, such as ``Chair'' and ``Shelf'', with another generation style like ``Bohemian'', and generate~(Fig.\ref{fig:scenario-2}(d)). Lucy can modify the generation style and the furniture pairings at the same time, another example is ``Chair
'' and ``Ottoman'' with ``Festive'' style, and view the generation results on the right panel~(Fig.\ref{fig:scenario-2}(e)). \\

\noindent \textbf{Final review.} For living room alone, apart from the aforementioned category combinations of ``Sofa'' with ``Lamp'', ``Chair'' and ``Shelf'', or ``Chair'' and ``Ottoman'', there are many other possible furniture pairings. And since viewing them one by one is very laborious, \ourname provides another use page called ``Final Gallery'' on the top menu bar~(Fig.\ref{fig:scenario-2}(g)). Lucy can enter the final review page by clicking the ``Final Gallery'' button and selecting a room type - ``Living room''. Then the system will fetch the generation results under all possible product combinations from the catalog for this room type. To have a user-friendly viewing panel, \ourname groups the generation by the product category pairs, and displays all the images under the same category pairing using a carousel, as shown in Fig.\ref{fig:scenario-2}(h). \\

\subsection{Automatic Evaluation}
In this section, we present evaluation results obtained using pre-trained models to automatically assess the quality of the generated ads. Given the goal of evaluating scalable generation objectively—and the constraints on resources and time for human evaluation—we relied on large language models (LLMs) and vision-language models (VLMs) as the most practical and effective alternatives. \\

\noindent \textbf{Ablations.} To demonstrate the effectiveness of each module in \ourname, namely product pairing module, layout generation module, and background generation module, we designed the following ablation studies to separately test them: (1) Remove product pairing~(A1), where we randomly sample products from the catalog and pair them together to generate an ad. (2) Remove product scaling~(A2), where we do not provide product metadata and also will not let the model extract product dimension information and use it as an input condition for subsequent generation. (3) Remove product placement~(A3), where instead of generating a description for product placement using VLM, we simply put the products side-by-side and perform subsequent generation. (4) Remove structural conditioning for background generation~(A4), where we turn off the ControlNets during background inpainting. For each ablation and our complete pipeline, we are sampling 40 generated ads to be evaluated, resulting in a total of 200 images.\\

\noindent \textbf{Metrics.} To evaluate the quality of generated ad images, we adopt both general-purpose and task-specific metrics. From the HEIM benchmark~\cite{10.5555/3666122.3669189}, we incorporate standard image quality measures, including CLIP-based image-text alignment~\cite{radford2021learning} and aesthetic scoring using the LAION Aesthetic Predictor.\footnote{\url{https://github.com/LAION-AI/aesthetic-predictor}}. Beyond these, we design a targeted evaluation framework for ad imagery by leveraging vision-language models~(VLMs) as automated evaluators. For each sampled generation, we supply the VLM with the original product images, the generation prompt, and the resulting ad image. The model is then prompted to rate the ad on a 1–5 scale across four dimensions:
(1) Product Authenticity — how realistic and convincing the ad is in showcasing the products;
(2) Visual Appeal — how visually attractive the ad image appears;
(3) Layout Quality — how well the image is composed in terms of balance, positioning, spacing, and the relative sizes of the two products;
(4) Theme Alignment — how well the background reflects the personalized theme specified in the generation prompt. To understand the variability in model-based evaluations, we compare results across two state-of-the-art VLMs: GPT-4o and GPT-4.5-preview. We further investigate the impact of structured prompting using Set-of-Mark (SoM)~\cite{yang2023setofmarkpromptingunleashesextraordinary}, assessing whether it improves rating consistency or interpretability for VLMs.\\

\noindent \textbf{Results.} We first present the automatic evaluation results using pre-trained models, namely the CLIP model for text-image alignment and LAION Aesthetic for overall visual attractiveness, in Table \ref{tab:ablation-1}. For text-image alignment, the differences between the ablations and the full pipeline are minimal, indicating that the impact of various components on this metric is relatively small. In contrast, the aesthetic scores seem to favor A1 and our approach (Ours) over the other configurations, particularly A3, which shows the lowest score. These general metrics, however, are not highly suitable or efficient in identifying qualified advertisements, as they fail to capture more nuanced aspects of ad quality that go beyond simple alignment and visual appeal.

\begin{table}[h]
  \centering
  \begin{tabular}{p{3cm}p{0.7cm}p{0.7cm}p{0.7cm}p{0.7cm}p{0.7cm}}
    \toprule
     & A1 & A2 & A3 & A4 & Ours\\
    \midrule
    CLIP($\uparrow$) & \underline{29.8} & 29.9 & 30.5 & \textbf{31.0} & 30.8 \\
    Aesthetic($\uparrow$) & 0.259 & 0.237 & \underline{0.212} & 0.230 &\textbf{0.267} \\
    \bottomrule
  \end{tabular}
  \caption{\textbf{Automatic ablation study using pre-trained models on \ourname.}}
  \label{tab:ablation-1}
\end{table}

Next, we report the quantitative evaluation results using VLMs. As shown in Table \ref{tab:ablation-authenticity}, we can conclude that while different model evaluators may favor certain configurations, A4 consistently shows a significant lack of product authenticity across all models. This is because, in this ablation, the structural constraints during background generation are removed, allowing the background to interfere with the product, thereby reducing the overall authenticity of the products being advertised.

\begin{table}[h]
  \centering
  \begin{tabular}{p{3cm}p{0.7cm}p{0.7cm}p{0.7cm}p{0.7cm}p{0.7cm}}
    \toprule
     & A1 & A2 & A3 & A4 & Ours\\
    \midrule
    GPT-4o & 4.410 & \textbf{4.6} & 4.355 & \underline{4.275} &  4.282 \\
    GPT-4o (SoM) & 4.410 & \textbf{4.700} & 4.484 & \underline{4.350} & 4.385 \\
    GPT-4.5-preview & 3.564 & \textbf{3.900} & 3.710 & \underline{3.150} & 3.641 \\
    GPT-4.5-preview (SoM) & 3.744 & 3.850 & 3.903 & \underline{3.175} & \textbf{3.923} \\
    \bottomrule
  \end{tabular}
  \caption{Ablation study on product authenticity.}
  \label{tab:ablation-authenticity}
\end{table}

We present the qualitative results on the visual appeal of the ablations in Table \ref{tab:ablation-visual}. Evaluation results from most VLMs show that A4 significantly outperforms all other pipelines in generating visually appealing images. This indicates that prioritizing product authenticity often comes at the cost of visual appeal. However, authenticity remains a key requirement in advertising applications, making the generation of high-quality ads a particularly challenging problem.

\begin{table}[h]
  \centering
  \begin{tabular}{p{3cm}p{0.7cm}p{0.7cm}p{0.7cm}p{0.7cm}p{0.7cm}}
    \toprule
     & A1 & A2 & A3 & A4 & Ours\\
    \midrule
    GPT-4o & 4.256 & \underline{4.125} & 4.194 & \textbf{4.375} & 4.282 \\
    GPT-4o (SoM) &  \textbf{4.077} & 4.075 & 4.032 &  4.075 & \underline{4.0256} \\
    GPT-4.5-preview & \underline{3.256} & 3.375 & 3.290 & \textbf{3.425} & 3.385 \\
    GPT-4.5-preview (SoM) & 3.000 & 3.150 & \underline{2.968} & \textbf{3.350} & 3.103 \\
    \bottomrule
  \end{tabular}
  \caption{Ablation study on visual appeal.}
  \label{tab:ablation-visual}
\end{table}

The ablation results for layout quality are reported in Table \ref{tab:ablation-layout}. From these results, we observe that different VLMs exhibit varying evaluation standards for layout composition. For instance, A2 is regarded as the worst layout configuration by GPT-4o, but conversely, it is evaluated as the highest quality layout by GPT-4.5-preview. This discrepancy highlights the subjectivity and model-dependent nature of layout evaluations, suggesting that different models prioritize distinct factors when assessing layout quality. These variations underscore the importance of considering multiple evaluation criteria when assessing the effectiveness of ad generation pipelines.

\begin{table}[h]
  \centering
  \begin{tabular}{p{3cm}p{0.7cm}p{0.7cm}p{0.7cm}p{0.7cm}p{0.7cm}}
    \toprule
     & A1 & A2 & A3 & A4 & Ours\\
    \midrule
    GPT-4o & 4.0 & \underline{3.775} & 3.903 & \textbf{4.05} & 3.846\\
    GPT-4o (SoM) & 3.949 & \underline{3.850} & \textbf{4.032} & 4.025 &  3.949\\
    GPT-4.5-preview & 2.462 & \textbf{2.475} & 2.419 & 2.425 & \underline{2.385} \\
    GPT-4.5-preview (SoM) & 2.436 & \textbf{2.675} & 2.452 & \textbf{2.675} & \underline{2.410} \\
    \bottomrule
  \end{tabular}
  \caption{Ablation study on layout quality.}
  \label{tab:ablation-layout}
\end{table}

Finally, we present the ablation results on theme alignment in Table \ref{tab:ablation-theme}. Similar to the findings in Table \ref{tab:ablation-visual}, A4 consistently outperforms the other configurations in terms of theme alignment. This suggests that removing structural constraints during background generation enhances the coherence of the theme in the generated advertisements. Furthermore, none of the ablations directly impacts the way the theme aligns with the background, which is why the observed differences in scores primarily reflect the variations introduced by using VLMs as a proxy for human evaluation. This underscores the limitations of relying solely on automated models for theme alignment, as they may not fully capture the nuanced judgment that human evaluators bring to assessing thematic coherence.

\begin{table}[h]
  \centering
  \begin{tabular}{p{3cm}p{0.7cm}p{0.7cm}p{0.7cm}p{0.7cm}p{0.7cm}}
    \toprule
     & A1 & A2 & A3 & A4 & Ours\\
    \midrule
    GPT-4o & 4.462 & \underline{4.375} & 4.419 & \textbf{4.650} &4.462 \\
    GPT-4o (SoM) & 4.205 & \textbf{4.275} & 4.258 & \textbf{4.275} & \underline{4.179} \\
    GPT-4.5-preview & \underline{3.5897} & 3.825 & 3.645 & 3.975 &  \textbf{4.103} \\
    GPT-4.5-preview (SoM) & \underline{3.436} &  3.825 & 3.387 &  \textbf{3.925} & 3.821 \\
    \bottomrule
  \end{tabular}
  \caption{Ablation study on theme alignment.}
  \label{tab:ablation-theme}
\end{table}

\begin{figure*}[h]
  \centering
  \includegraphics[width=\linewidth]{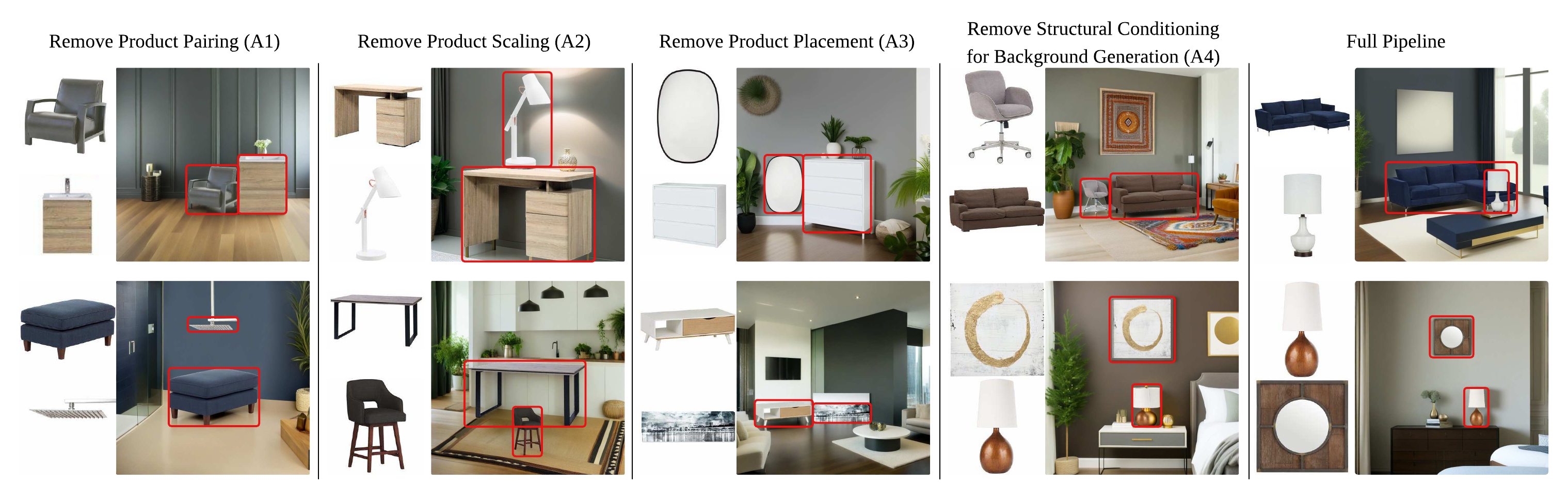}
  \caption{Qualitative results of ablations on sampled images. Removing the product pairing module (A1) leads to functionally mismatched products, removing the size estimation module (A2) results in unrealistic product sizes, removing the layout planning module (A3) causes unsatisfactory side-by-side alignment, and removing structural constraints during background generation (A4) results in inappropriate product appendages, violating authenticity.}
  \label{fig:qualitative-ablation}
\end{figure*}

Additionally, we provide qualitative results on sampled images with each ablation in Fig. \ref{fig:qualitative-ablation}. From these images, we can draw the following conclusions: removing the product pairing module (A1) leads to the combination of functionally distinct products, such as a sofa and a shower, which makes the image unrealistic. Removing the size estimation module (A2) results in products with unrealistic relative sizes. Removing the layout planning module (A3) causes the products to be aligned side-by-side, which can be visually unsatisfactory. Finally, removing the structural constraints during background generation (A4) leads to products having inappropriate appendages, such as adding legs to sofas or extending lamps, significantly violating the authenticity required for advertisements.

\section{Discussion}
\subsection{Findings} %
\noindent \textbf{Trade-off between product authenticity and visual appeal.} While removing structural constraints during background generation results in more visually appealing images by allowing for greater flexibility in design and aesthetic choices, it comes at the cost of compromising the authenticity of the product pairings. In advertising, authenticity is a crucial element, as consumers expect product representations that are not only visually attractive but also believable and consistent with real-world scenarios. When structural constraints are relaxed, the system may combine products in unrealistic ways, which diminishes the credibility of the ad. This trade-off underscores the inherent challenge in ad generation: design systems must find a delicate balance between creating visually engaging content and maintaining the integrity of the products being advertised. The challenge lies in achieving a design that appeals to the viewer while preserving the authenticity and relatability that are essential for successful advertising. As such, this balancing act between aesthetics and authenticity is a key obstacle when attempting to generate high-quality advertisements that resonate with audiences and meet brand expectations.\\

\noindent \textbf{Multi-process generation pipeline.} The results demonstrate that our multi-process generation pipeline not only produces higher-quality ads but also offers users the flexibility to adjust the generation details according to their preferences. This dual benefit highlights the effectiveness of our approach in both optimizing the automated ad generation process and accommodating user input. By providing this level of flexibility, our system supports a design pattern that incorporates two levels of granularity: end-to-end generation for seamless, automated ad creation, and low-level controls that allow users to fine-tune specific aspects of the generated content. This design pattern caters to different user needs—those seeking quick, fully automated results, as well as those requiring more personalized, customized adjustments. Such a hybrid approach is particularly valuable in applications where user-specific customization is crucial, ensuring that the generated ads not only meet high-quality standards but also align with individual or brand-specific preferences.\\

\noindent \textbf{Implications for practices.} The variation in results across different VLMs indicates that current general-purpose models struggle to effectively evaluate specific ad images. While these models provide useful insights, their ability to assess the unique qualities required for high-quality advertisements—such as authenticity, appeal, and context—remains limited. For instance, general VLMs might prioritize factors like text-image alignment or overall visual attractiveness, but they fail to automatically capture the subtle nuances that are critical in advertising, such as product consistency. As a result, their evaluations may not always align with human judgment, which is inherently more attuned to the emotional and contextual aspects of ads. This discrepancy highlights the ongoing need for human evaluations in the ad generation process. Human evaluators can provide nuanced feedback that reflects real-world preferences and standards, which current models still cannot replicate. Thus, while VLMs can serve as useful tools, integrating human assessments will likely remain essential for achieving optimal results in ad creation and evaluation. \\

\subsection{Limitation}
\noindent \textbf{Generalization to more diverse products.} Currently, \ourname is limited to furniture products. However, e-commerce advertising encompasses a wide range of other product categories—such as clothing, electronics, home decor, and personal care items—that could similarly benefit from complementary and visually compelling ad generation. Extending the system to support these domains would require adapting the layout reasoning, visual compatibility filters, and background generation strategies to account for different product characteristics. Enabling such generalization would significantly broaden the applicability and impact of the approach across the e-commerce ecosystem. \\

\noindent \textbf{Generation performance.} Although \ourname is capable of generating well-composed ad images with appropriate positioning, spacing, and realistic relative sizing, failure cases still occur. These issues can be attributed to the limitations of general-purpose VLMs and inpainting models, which often lack domain-specific knowledge in furniture advertising. Incorporating retrieval-augmented, few-shot learning for VLMs, as well as fine-tuning background generation models with domain-relevant data, could offer promising directions to address these limitations and enhance generation quality.\\

\noindent \textbf{Generation speed.} In \ourname, generating a single image takes approximately 10 seconds on an NVIDIA A100 GPU. Scaling this process to tens or even hundreds of thousands of images can require several days, significantly limiting the efficiency of iterative human feedback during large-scale generation. This latency poses a bottleneck for rapid monitoring and refinement. Potential solutions include enabling parallel generation across multiple GPUs or applying quantization techniques to trade off image quality for faster generation speed. \\

\noindent \textbf{System design.} Currently, the system only stores user adjustments on a per-instance basis. To support more efficient monitoring and refinement at scale, it would be beneficial to enable batch operations that allow users to apply adjustments across multiple generations. For instance, when a user modifies the layout between a specific coffee table and sofa pair, the system could provide an option to propagate these changes to other visually similar pairs. This capability would help improve overall generation quality while reducing repetitive manual effort. \\

\section{Future Work}
\subsection{3D Products}
In \ourname, the generation pipeline is currently constrained to 2D representations due to the need for authentically showcasing products. Most businesses provide 2D product images in their catalogs, while 3D models are rarely available. As a result, we are limited to using the exact 2D product images captured from a specific viewpoint, without applying novel view synthesis—which could compromise product authenticity. However, if 3D product models were available, the system’s capabilities could be significantly expanded. For instance, it would be possible to construct 3D room scenes, place products within a spatial context, and render them from multiple perspectives. This would enable more dynamic ad creation, allowing for novel viewpoints, realistic lighting, shadows, and reflection effects—ultimately enhancing the visual richness and flexibility of the generated content.

\subsection{Video generation}
Another emerging need in advertising is to create immersive shopping experiences for users. While images are effective at conveying visual information intuitively, they lack the interactivity and engagement offered by videos. Extending the output format from static images to video could therefore provide significant value. For example, generating 360-degree room renders, incorporating simple camera movements such as zooming in or out, or animating background elements could enhance the sense of presence and immersion. Such dynamic content would not only improve user engagement but also offer a more compelling and informative presentation of the products.

\subsection{Automatic filtering}
The role of users in \ourname is two-fold: they supervise and intervene in the generation process when necessary, and they also serve as the final evaluators before images are deployed to production. Although the system is designed to reduce manual effort in reviewing large batches of generations, the scale—often involving thousands of images—still requires substantial time and attention. To further streamline this workflow and support scalable operations, automatic filtering mechanisms can be introduced to pre-screen results before they are shown to users. Potential approaches include training task-specific models to identify low-quality outputs, or applying unsupervised clustering to group similar generations for more efficient review. An additional enhancement could involve grouping generations by the confidence level of the automatic filter, allowing users to focus their attention primarily on lower-confidence results. This enables a more efficient collaboration between AI and human reviewers, preserving quality while reducing human workload.

\section{Conclusion}
This paper presents \ourname, a scalable ad image generation system that enables users to create visually appealing, high-quality lifestyle images from large-scale product catalogs. By combining vision-language models, layout reasoning, and interactive refinement tools, \ourname supports both automated generation and user-guided intervention. Through scenario-based exploration and module evaluations, we demonstrated how \ourname facilitates efficient creation of personalized ad content across diverse room types and themes. %
Looking ahead, future work could explore generalization to other product categories (e.g., clothing or appliances), integration of 3D models for immersive content, and the use of video outputs to support more engaging shopping experiences. We also see potential in further optimizing human-AI collaboration, such as confidence-based filtering and batch editing, to streamline the review process and support truly scalable content production.

\bibliographystyle{ACM-Reference-Format}
\bibliography{Sections/main}

\clearpage
\newpage
\newpage

\appendix
\section{Additional Examples of \ourname Generated Ads}
\begin{figure*}[h]
  \centering
  \includegraphics[width=0.8\linewidth]{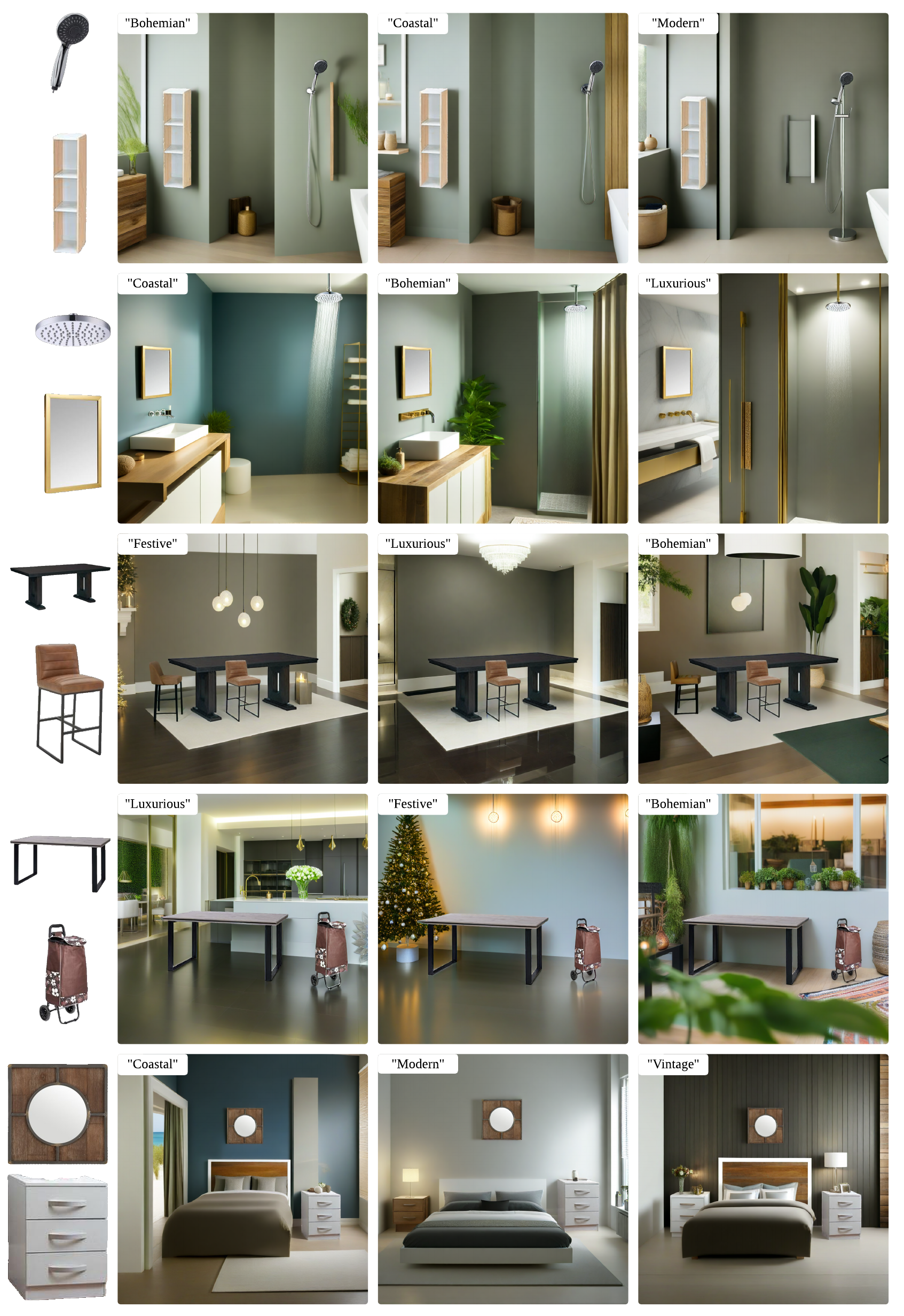}
  \caption{Examples of other generated advertisements.}
  \label{fig:gallery-1}
\end{figure*}

\begin{figure*}[h]
  \centering
  \includegraphics[width=0.8\linewidth]{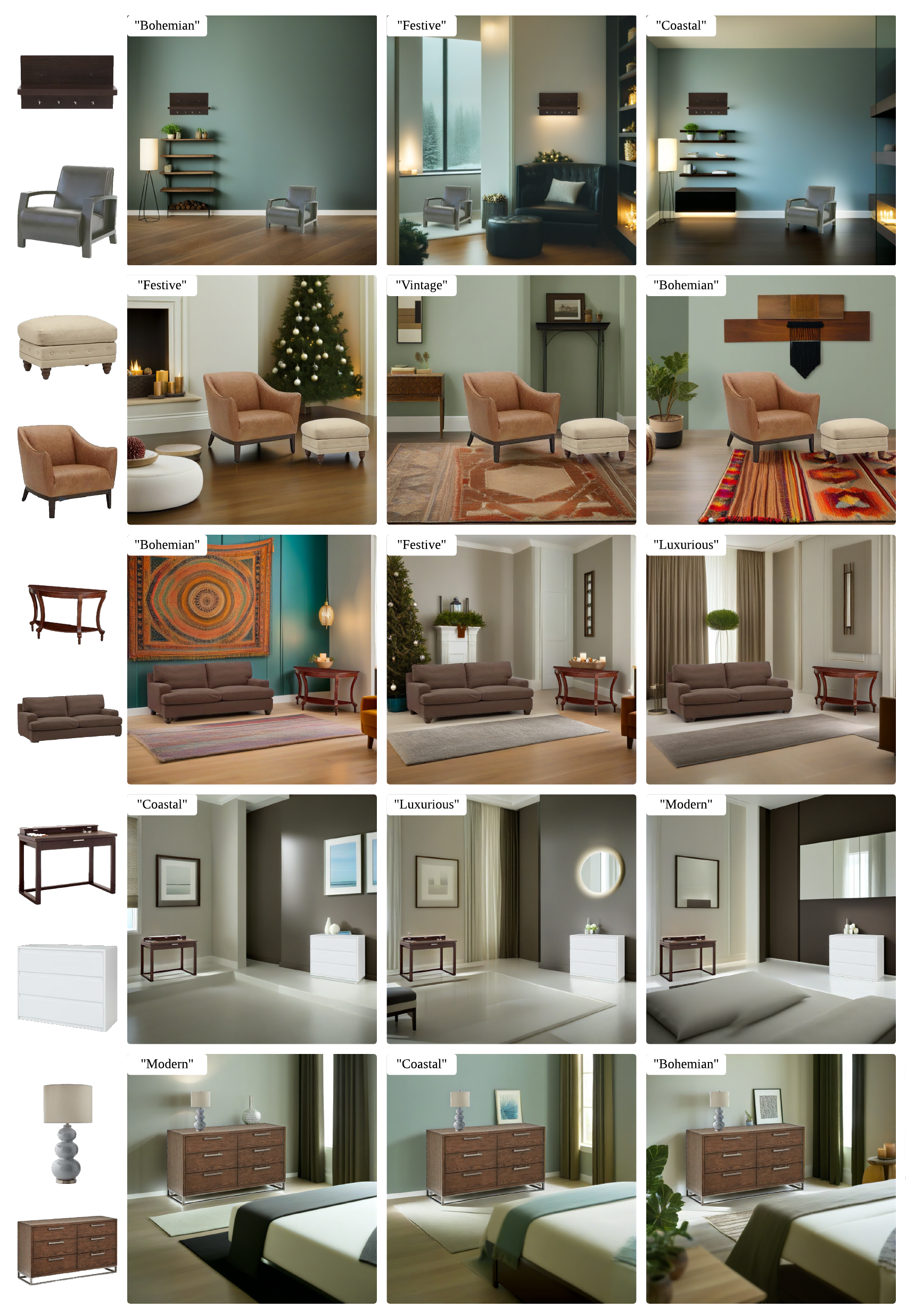}
  \caption{Examples of other generated advertisements.}
  \label{fig:gallery-2}
\end{figure*}

We provide more examples of advertisement generated using \ourname in Fig.~\ref{fig:gallery-1} and Fig.~\ref{fig:gallery-2}.

\section{Implementation Details}
\subsection{Prompt Templates in \ourname}
Here we provide the prompt template we used for the vision-language models in \ourname's image generation pipeline. \\

\noindent \textbf{Room Type Selection and Product Information Retrieval } \\
Prompt: ``You are an Advertising Marketing Expert. You are given a product images and product title information: {title}. You will generate a short product description, extract product sizes, and decide which room type this product should be placed. Let's think step by step:
    1. Does the product showcased in the image falls under the category of \{category\}? Answer Yes or No.
    2. Describe the product in the image. Please be sure to summarize the product into a short product description within three words, such as "pink perfume bottle".
    3. Extract the product dimensions from the product title information. Answer by giving only: length x width x height in cm
    4. Choose one of the room types from "bedroom, living room, kitchen, bathroom" where this product should be placed. Answer by giving only the room type.
Return the answer to each question in a JSON format:\{\{"question number" : "answer"\}\} ''\\

The \{category\} blank is filled with the corresponding product category. And the final input format for the vision-language model is as follows:
\begin{lstlisting}
messages = [
{
    "role": "user",
    "content": [
        {
        "type": "text",
        "text": prompt
        },
        {
        "type": "image_url",
        "image_url": {
            "url": f"data:image/png;base64,{base64_image}",
            "detail": "high"
        }
        }
    ]
}]
\end{lstlisting}

\noindent \textbf{Layout and Background Description} \\
Prompt: ``You are an Advertising Marketing Expert. You are given two product images, and will generate text for these products to show as product photography, which contains product descriptions and a overall image description. Let's think step by step:
    1. Describe the product in the first image. Please be sure to summarize the product into a short product description within three words, such as "pink perfume bottle".
    2. Describe the product in the second image. Please be sure to summarize the product into a short product description within three words, such as "pink perfume bottle".
    3. Come up with a layout-prompt within 30 words that describes how the layout of the \{room type\} looks like containing the products. The layout must be suitable to showcase the two products.
    4. Given the theme as \{generation style\}, come up with an overall photo description that combines the short product descriptions, the layout description, and the background description within 40 words.
Return the answer to each question in a JSON format:\{\{"question number" : "answer"\}\}''

The \{room type\} and \{generation style\} blanks are filled with the corresponding room type and generation style for the current generation. And the final input format for the vision-language model is as follows:

\begin{lstlisting}
messages = [
{
    "role": "user",
    "content": [
        {
        "type": "text",
        "text": prompt
        },
        {
        "type": "image_url",
        "image_url": {
            "url": f"data:image/png;base64,{base64_image1}",
            "detail": "high"}
        },
        {
        "type": "image_url",
        "image_url": {
            "url": f"data:image/png;base64,{base64_image2}",
            "detail": "high"}
        }
    ]
}]
\end{lstlisting}

\noindent \textbf{Layout Generation} \\
System Prompt: ``Instruction: Given a sentence prompt containing products that will be used to generate an image, plan the layout of the products. The generated layout should follow the CSS style, where each line starts with the object description and is followed by its absolute position and the layering. The layering number starts from 0, with 0 representing foreground and 1 representing background. Formally, each line should be like "object \{\{width: ?px; height: ?px; left: ?px; top: ?px; layer: ?\}\}". The image is 1024px wide and 1024px high, with the floor line on 768px from top as your reference. Therefore, all properties of the positions should not exceed 1024px, The addition of left and width, and the addition of top and height should not exceed 1024px.''\\

\noindent User Text Prompt: ``A \{product-1\} with width-to-height ratio of \{aspect-ratio-1\}, and \{aspect-ratio-2\} with width-to-height ratio of \{aspect-ratio-2\}. The canvas represents a room of size (in cm) 500 x 400 x 200. The layout should follow \{layout-prompt\}. Return the layout CSS with two lines, with first line \{product-1\} and second line \{product-2\}." \\

The \{product-1\}, \{product-1\}, and {layout-prompt\} blanks are filled with the product summary and layout descriptions generated from the VLM in the prior steps. The \{aspect-ratio-1\} and \{aspect-ratio-2\} blanks are filled with the aspect ratio of the original product images retrieved from the catalog. And the final input format for the vision-language model is as follows:

\begin{lstlisting}
message_list = [
    {
        'role': 'system', 
        'content': system_prompt
    },
    {   
        "role": "user", 
        "content": [
        {
            "type": "text",
            "text": user_text_prompt,
        },
        {
            "type": "image_url",
            "image_url": {
                "url": f"data:image/png;base64,{base64_image1}",
                "detail": "high"
            }
        },
        {
            "type": "image_url",
            "image_url": {
                "url": f"data:image/png;base64,{base64_image2}",
                "detail": "high"}
        }]
    }
]
\end{lstlisting}

\subsection{Prompt Templates for Evaluations}
Here we provide the prompt template we used for using vision-language models to evaluate \ourname.\\

\noindent \textbf{Product Authenticity} \\
System prompt: ``Instruction: Given a text prompt and two product images, please evaluate the ad image and give a score from 1 to 5. The score should be based on the how authentic the featured products in the ad image are to the original product images.The score should be a number between 1 and 5, where 1 means the two products shown in the ads are very different from their original images, and 5 means the ad image is exactly featuring the two products. Please provide a detailed explanation of the score you give. Return the score and explanation in a JSON format, with the score as a number and the explanation as a string. The JSON format should be like this: {"score": , "explanation": } Please do not include any other text or information in the response.''\\

\noindent \textbf{Visual Appeal} \\
System Prompt: ``Instruction: Given a text prompt and two product images, please evaluate the ad image and give a score from 1 to 5. The score should be based on how visually appealing is the image as an advertisement for the two products. The score should be a number between 1 and 5, where 1 means the ad image is unappealing and 5 means the ad image is very appealing. Please provide a detailed explanation of the score you give. Return the score and explanation in a JSON format, with the score as a number and the explanation as a string. The JSON format should be like this: {"score": , "explanation": }. Please do not include any other text or information in the response.''\\

\noindent \textbf{Layout Quality} \\
System Prompt: ``Instruction: Given a text prompt and two product images, please evaluate the scene image and give a score from 1 to 5. The score should be based on the how well-composed is the scene image as an advertisement layout (balance, positioning, spacing, realistic relative size between the two products. The score should be a number between 1 and 5, where 1 means the ad is very ill-composed, and 5 means the ad image is extremely well composed. Please provide a detailed explanation of the score you give. Return the score and explanation in a JSON format, with the score as a number and the explanation as a string. The JSON format should be like this: {"score": , "explanation": }. Please do not include any other text or information in the response.''\\

\noindent \textbf{Theme Alignment} \\
System Prompt: ``Instruction: Given a text prompt and two product images, please evaluate the scene image and give a score from 1 to 5. The score should be based on to what extent does the background align with the personalized theme in the generation prompt. The score should be a number between 1 and 5, where 1 means ad image is very poorly aligned with the theme, and 5 means the ad image is very well aligned with the theme. Please provide a detailed explanation of the score you give. Return the score and explanation in a JSON format, with the score as a number and the explanation as a string. The JSON format should be like this: {"score": , "explanation": } Please do not include any other text or information in the response.''\\

If Set-of-Mark is used, then we add ``The two products are highlighted separately with a red and a green bounding box in the ad image.'' after the first sentence in each system prompt.

The complete message to the VLM consists of:
\begin{lstlisting}
messages = [
{
    "role": "user",
    "content": [
        {
        "type": "text",
        "text": text_prompt
        },
        {
        "type": "image_url",
        "image_url": {
            "url": f"data:image/png;base64,{base64_image1}",
            "detail": "high"
        }
        },
        {
        "type": "image_url",
        "image_url": {
            "url": f"data:image/png;base64,{base64_image2}",
            "detail": "high"
        }
        },
        {
        "type": "image_url",
        "image_url": {
            "url": f"data:image/png;base64,{base64_ad}",
            "detail": "high"
        }
        }
    ]
}]
\end{lstlisting}

\subsection{Raw Data Format of the Dataset}
Here we include an example of the raw data extracted from the Amazon-Berkeley-Object dataset, which is the data input format to \ourname.

\begin{lstlisting}
example_info = {
    'brand': [
        {'language_tag': 'en_IN', 'value': 'Amazon Brand - Solimo'}
    ], 
    'bullet_point': [
        {'language_tag': 'en_IN', 'value': 'Snug fit for Samsung\xa0Galaxy\xa0J2\xa0Ace, with perfect cut-outs for volume buttons, audio and charging ports'}, {'language_tag': 'en_IN', 'value': 'Compatible with Samsung\xa0Galaxy\xa0J2\xa0Ace'}, 
        {'language_tag': 'en_IN', 'value': 'Easy to put & take off with perfect cutouts for volume buttons, audio & charging ports.'}, 
        {'language_tag': 'en_IN', 'value': 'Stylish design and appearance, express your unique personality.'}, {'language_tag': 'en_IN', 'value': 'Extreme precision design allows easy access to all buttons and ports while featuring raised bezel to life screen and camera off flat surface.'}
    ], 
    'color': [
        {'language_tag': 'en_IN', 'standardized_values': ['multi-colored'], 'value': 'Multicolor'}
    ], 
    'item_id': 'B0857LSVB7', 
    'item_name': [
        {'language_tag': 'en_IN', 'value': 'Amazon Brand - Solimo Designer Lion UV Printed Soft Back Case Mobile Cover for Samsung\xa0Galaxy\xa0J2\xa0Ace'}
    ], 
    'item_weight': [
        {'normalized_value': {'unit': 'pounds', 'value': 0.110231131}, 'unit': 'grams', 'value': 50}
    ], 
    'material': [
        {'language_tag': 'en_IN', 'value': 'Silicon'}
    ],
    'model_name': [
        {'language_tag': 'en_IN', 'value': 'Samsung\xa0Galaxy\xa0J2\xa0Ace'}
    ], 
    'model_number': [{'value': 'UV10392-SL40350'}], 'product_type': [{'value': 'CELLULAR_PHONE_CASE'}], 'main_image_id': '81-DuD5XzmL', 
    'other_image_id': ['61+woWTqkwL', '61SE4RTPjdL'], 'item_keywords': [
        {'language_tag': 'en_IN', 'value': 'Back Cover'},
        {'language_tag': 'en_IN', 'value': 'Designer Case'}, 
        {'language_tag': 'en_IN', 'value': 'Designer Lion Mobile Cover'}, 
        {'language_tag': 'en_IN', 'value': 'Flexible Case'},
        {'language_tag': 'en_IN', 'value': 'Printed Cover'},
        {'language_tag': 'en_IN', 'value': 'Samsung\xa0Galaxy\xa0J2\xa0Ace Case'}, 
        {'language_tag': 'en_IN', 'value': 'Silicone Case'},
        {'language_tag': 'en_IN', 'value': 'Soft TPU'}, 
        {'language_tag': 'en_IN', 'value': 'cases and covers'}, 
        {'language_tag': 'en_IN', 'value': 'fashion case'}, 
        {'language_tag': 'en_IN', 'value': 'mobile Cover'}], 
    'country': 'IN', 
    'marketplace': 'Amazon', 
    'domain_name': 'amazon.in', 
    'node': [
        {'node_id': 12538061031, 'node_name': '/Categories/Mobiles & Accessories/Mobile Accessories/Maintenance, Upkeep & Repairs/Replacement Parts/Back Covers'}, 
        {'node_id': 12710103031, 'node_name': '/Categories/Mobiles & Accessories/Mobile Accessories/Cases & Covers/Back & Bumper Cases'}
    ]
}
\end{lstlisting}

\end{document}